\def\year{2017}\relax
\PassOptionsToPackage{usenames,dvipsnames}{xcolor}
\documentclass[EPiC,withtimes]{easychair}
\usepackage{doc}

\usepackage{times}
\usepackage{algorithm,algorithmicx}
\usepackage[noend]{algpseudocode}

\newcommand*\Let[2]{\State #1 $\gets$ #2}
\usepackage{hhline}
\usepackage{paralist}
\usepackage{stmaryrd}
\usepackage{relsize}
\usepackage{amsthm}

\usepackage{import}

\usepackage{ifthen}
\usepackage{url}
\usepackage{amssymb}
\usepackage{amsmath}
\usepackage{import}
\usepackage{xparse}
\usepackage{amsmath}
\usepackage[usenames,dvipsnames]{xcolor}
\usepackage{xspace}
\usepackage{datatool}
\usepackage{glossaries}

\providecommand\m[1]{\ensuremath{#1}\xspace}
\renewcommand{\m}[1]{\ensuremath{#1}\xspace}
\newcommand{\trval}[1]{\m{\mathbf{#1}}}

	\newcommand{\lrule}{\leftarrow}
	\newcommand{\cause}{\stackrel{c}{\lrule}}
	
	\newcommand{\ltrue}{\trval{t}}
	\newcommand{\lfalse}{\trval{f}}
	\newcommand{\lunkn}{\trval{u}}
	\newcommand{\lincon}{\trval{i}}

	\newcommand{\voc}{\m{\Sigma}}

	\newcommand{\struct}{\m{I}}
	
	\newcommand{\I}{\m{\mathcal{I}}}

	\newcommand{\PP}{\m{\mathcal{P}}}

	\NewDocumentCommand\inter{g+g}{%
	  \IfNoValueTF{#1}
	    {\struct}
	    {\m{#1^{#2}}}}

	\newcommand{\ddd}{\m{\overline{d}}}

	\newcommand{\bool}{\m{\mathbb{B}}}
	
	\newcommand{\nat}{\m{\mathbb{N}}}
	
	\renewcommand{\int}{\m{\mathbb{Z}}}

	\newcommand{\leqp}{\m{\leq_p}}
	\newcommand{\geqp}{\m{\geq_p}}

	\DeclareMathOperator\glb{glb}
	\DeclareMathOperator\lub{lub}
	\DeclareMathOperator\lfp{lfp}

	\NewDocumentCommand\subs{g+g}{%
	  \IfNoValueTF{#1}
	    {\m{/}}
	    {\m{#1/ #2}}}

\newcommand{\ouracronym}[3]{%
	\newacronym{#1}{#2}{#3}
	\expandafter\newcommand\csname #1\endcsname{\gls{#1}\xspace}%
}
	\ouracronym{FO}{FO}{first-order logic}
	\ouracronym{PC}{PC}{propositional calculus}
	\ouracronym{MX}{MX}{Model Expansion}
	\ouracronym{MO}{MO}{Model Optimization}
	\ouracronym{ASP}{ASP}{Answer Set Programming}
	\ouracronym{TP}{TP}{Theorem Proving}
	\ouracronym{LP}{LP}{Logic Programming}
	\ouracronym{CP}{CP}{Constraint Programming}
	\ouracronym{FP}{FP}{Functional Programming}
	\ouracronym{KR}{KR}{Knowledge Representation}
	\ouracronym{CSP}{CSP}{Constraint Satisfaction Problem}
	\ouracronym{SMT}{SMT}{SAT Modulo Theories}
	\ouracronym{KBS}{KBS}{Knowledge Base System}
	\ouracronym{NNF}{NNF}{Negation Normal Form}
	\ouracronym{FNNF}{FNNF}{Flat Negation Normal Form}
	\ouracronym{DefNNF}{DefNNF}{Definition Negation Normal Form}
	\ouracronym{DEFNF}{DEFNF}{Definition Normal Form}
	\ouracronym{CDCL}{CDCL}{Conflict-Driven Clause-Learning}
	\ouracronym{WFS}{WFS}{Well-Founded Semantics}
	\ouracronym{LCG}{LCG}{Lazy Clause Generation}
	\ouracronym{AEL}{AEL}{Autoepistemic Logic}
	\ouracronym{OEL}{OEL}{Ordered Epistemic Logic}
	\ouracronym{AFT}{AFT}{Approximation Fixpoint Theory}

	\makeatletter
	\def\ifenv#1{
	\def\@tempa{#1}%
	\def\@ttempa{#1*}%
	\ifx\@tempa\@currenvir
	\expandafter\@firstoftwo
	\else
	\expandafter\@secondoftwo
	\fi
	}
	\makeatother

	\newcommand{\ddrule}[4]{\ensuremath{#1 \leftarrow #2 & \{#3\} & #4}}
	\newcommand{\drule}[2]{\ensuremath{#1 & \leftarrow & #2}}

	\newcommand{\darule}[4]{\ensuremath{#1 \leftarrow #2 & \{#3\} & #4}}
	\newcommand{\arule}[2]{\ensuremath{#1 \, &\leftarrow \, #2}}

	\newcommand{\LNDRule}[2]{
	\ifenv{array}
	{\drule{#1}{#2}}
	{ \ifenv{align}
		{\arule{#1}{#2}}
		{\ifenv{align*}
		{\arule{#1}{#2}}
		{ERROR: using LDRule in unsupported environment: \@currenvir}
		}
	}
	}

	\newcommand{\LDRule}[4]{
	\ifenv{array}
	{\ddrule{#1}{#2}{#3}{#4}}
	{ \ifenv{align}
		{\darule{#1}{#2}{#3}{#4}}
		{\ifenv{align*}
		{\darule{#1}{#2}{#3}{#4}}
		{ERROR: using LDRule in unsupported environment: \@currenvir}
		}
	}
	}

	\NewDocumentCommand\LRule{m+g+g+g}{%
		\IfNoValueTF{#2}%
		{#1.&}{%
		\IfNoValueTF{#3}
		{\LNDRule{#1}{#2.}}
		{\LDRule{#1}{#2.}{#3}{#4}}%
		}
	}

	\NewDocumentCommand\CLRule{m+g}{%
	\ifenv{array}
	{\cdrule{#1}{#2}}
	{ \ifenv{align}
		{\carule{#1}{#2}}
		{\ifenv{align*}
			{\carule{#1}{#2}}
			{ERROR: using CLRule in unsupported environment: \@currenvir}
		}
	}
	}

	\NewDocumentCommand\carule{m+g}{%
		\IfNoValueTF{#2}
			{\ensuremath{#1.}}
			{\ensuremath{#1 \, &\cause \, #2}}}
	\NewDocumentCommand\cdrule{m+g}{%
		\IfNoValueTF{#2}
			{\ensuremath{#1.}}
			{\ensuremath{#1 & \cause & #2}}}

	\newcommand{\algrule}[4]{
	\hbox{{#1}:}& 
	\quad #2 ~\longrightarrow~ #3 
	\hbox{~ if } #4\\
	}

	\newcommand{\AlgoRule}[4]{
	\ifenv{array}
	{\algrule{#1}{#2}{#3}{#4}}
		{ERROR: using AlgoRule in unsupported environment: \@currenvir}
	}

	\newcommand{\ignore}[1]{}

	\newboolean{nocomments}
	\setboolean{nocomments}{false}

	\newboolean{commentmargin}
	\setboolean{commentmargin}{true}

	\newcommand{\namedcomment}[3]{%
		\ifthenelse{\boolean{nocomments}}%
		{}
		{
			\ifthenelse{\boolean{commentmargin}}%
				{ {\color{#3} \marginpar{\color{#3}\sc #2}#1}  }
				{  {\color{#3} {\sc #2}: #1}  }
		}%
	}
	\newcommand{\mnamedcomment}[3]{\ifthenelse{\boolean{nocomments}}{}{{\marginpar{ \color{#3}{\sc #2}:#1}}}}

	\newcommand{\todo}[1]{\namedcomment{#1}{TODO}{blue}}

	\usepackage{soul}

\usepackage{etoolbox}

\newcommand\setcitation[2]{%
  \csdef{mycommoncitation#1}{#2}}
\newcommand\getcitation[1]{%
  \csuse{mycommoncitation#1}}

\setcitation{IDP}{WarrenBook/DeCatBBD14}
\setcitation{idp}{WarrenBook/DeCatBBD14}
\setcitation{fodot}{tocl/DeneckerT08}
\setcitation{CP}{Apt03}
\setcitation{cp}{Apt03}
\setcitation{EZCSP}{lpnmr/Balduccini11}
\setcitation{KR}{Baral:2003}
\setcitation{ASPComp2}{lpnmr/DeneckerVBGT09}
\setcitation{ASPComp3}{journals/tplp/CalimeriIR14}
\setcitation{ASPComp4}{conf/lpnmr/AlvianoCCDDIKKOPPRRSSSWX13}
\setcitation{CPSupport}{ictai/DeCat13}
\setcitation{CPsupport}{ictai/DeCat13}
\setcitation{functionDetection}{iclp/DeCatB13}
\setcitation{fodot2asp}{DeneckerLTV12} 
\setcitation{Tarskian}{DeneckerLTV12} 
\setcitation{Inca}{iclp/DrescherW12}
\setcitation{csp2asp}{ijcai/DrescherW11}
\setcitation{DPLLT}{cav/GanzingerHNOT04}
\setcitation{AspInPractice}{synthesis/2012Gebser}
\setcitation{clasp}{ai/GebserKS12}
\setcitation{oclingo}{kr/GebserGKOSS12}
\setcitation{gringo}{lpnmr/GebserST07}
\setcitation{cmodels}{aaai/GiunchigliaLM04}
\setcitation{inputster}{tplp/Jansen13}
\setcitation{DLV}{tocl/LeonePFEGPS06}
\setcitation{LearningPaper}{TPLP/BruynoogheBBDDJLRDV} 
\setcitation{clog}{iclp/BogaertsVDV14} 
\setcitation{foc}{iclp/BogaertsVDV14}
\setcitation{FOC}{iclp/BogaertsVDV14}
\setcitation{inferenceClog}{ecai/BogaertsVDV14}
\setcitation{examplesClog}{nmr/BogaertsVDV14b} 
\setcitation{AFT}{DeneckerBV12}
\setcitation{KBS}{iclp/DeneckerV08}
\setcitation{KBS-invitedtalk}{jelia/Denecker16}
\setcitation{KBPE}{inap/DePooterWD11}
\setcitation{lazyGrounding}{jair/CatDBS15} 
\setcitation{lazygrounding}{jair/CatDBS15} 
\setcitation{ASP}{marek99stable}
\setcitation{satid}{sat/MarienWDB08}
\setcitation{lazyclausegeneration}{constraints/OhrimenkoSC09}
\setcitation{FP}{ACMCS/Hudak89}
\setcitation{GroundingWithBounds}{jair/WittocxMD10}
\setcitation{SAT}{faia/SilvaLM09}
\setcitation{LTC}{iclp/Bogaerts14}
\setcitation{SPSAT}{ictai/DevriendtBMDD12}
\setcitation{BreakID}{sat/DevriendtBBD16}
\setcitation{breakid}{sat/DevriendtBBD16}
\setcitation{LCG}{stuckeyLCG}
\setcitation{MiniZinc}{conf/cp/NethercoteSBBDT07}
\setcitation{minizinc}{conf/cp/NethercoteSBBDT07}
\setcitation{amadini}{cpaior/AmadiniGM13}
\setcitation{bootstrapping}{ngc/BogaertsJDJBD16}
\setcitation{Bootstrapping}{ngc/BogaertsJDJBD16}
\setcitation{GroundedFixpoints}{ai/BogaertsVD15}
\setcitation{PartialGroundedFixpoints}{ijcai/BogaertsVD15}
\setcitation{LogicBlox}{datalog/GreenAK12}
\setcitation{proB}{journals/sttt/LeuschelB08}
\setcitation{NaturalInductions}{KR/DeneckerV14} 
\setcitation{LP}{jacm/EmdenK76}
\setcitation{SMT}{faia/BarrettSST09}
\setcitation{AF}{ai/Dung95}
\setcitation{ADF}{kr/BrewkaW10}
\setcitation{af}{ai/Dung95}
\setcitation{adf}{kr/BrewkaW10}
\setcitation{ADFRevisited}{ijcai/BrewkaSEWW13}
\setcitation{adfrevisited}{ijcai/BrewkaSEWW13}
\setcitation{DefaultLogic}{ai/Reiter80}
\setcitation{DL}{ai/Reiter80}
\setcitation{AEL}{mo85}
\setcitation{minisat}{sat/EenS03}
\setcitation{completion}{adbt/Clark78}
\setcitation{wasp}{lpnmr/AlvianoDFLR13}
\setcitation{minisatid}{ictai/DeCat13}
\setcitation{lcg}{stuckeyLCG}
\setcitation{CEGAR}{jacm/ClarkeGJLV03}
\setcitation{cegar}{jacm/ClarkeGJLV03}
\setcitation{CuttingPlane}{or/DantzigFJ54}
\setcitation{kodkod}{tacas/TorlakJ07}
\setcitation{cdcl}{Marques-SilvaS99}
\setcitation{CDCL}{Marques-SilvaS99}
\setcitation{relevance}{ijcai/JansenBDJD16}
\setcitation{relevance-implementation}{aspocp/JansenBDJD16}
\setcitation{WFS}{GelderRS91}
\setcitation{wfs}{GelderRS91}
\setcitation{UnfoundedSet}{GelderRS91}
\setcitation{stablesemantics}{iclp/GelfondL88}
\setcitation{StableSemantics}{iclp/GelfondL88}
\setcitation{shatter}{Shatter}
\setcitation{sbass}{drtiwa11a}
\setcitation{lparsemanual}{url:lparse_manual}
\setcitation{AIC}{ppdp/FlescaGZ04}
\setcitation{templates}{tplp/DassevilleHJD15}
\setcitation{templates2}{iclp/DassevilleHBJD16}
\setcitation{sat-to-sat}{aaai/JanhunenTT16}
\setcitation{sat-to-sat-qbf}{bnp/BogaertsJT16}
\setcitation{sat-to-sat-QBF}{bnp/BogaertsJT16}
\setcitation{sat-to-sat-SO}{kr/BogaertsJT16}
\setcitation{XSB}{SwiW12}
\setcitation{KCmap}{jair/DarwicheM02}
\setcitation{TLA}{DBLP:books/aw/Lamport2002}
\setcitation{EventB}{BookAbrial2010}
\setcitation{MX}{MitchellTHM06}
\setcitation{MIP}{Sierksma96}
\setcitation{perefectmodel}{minker88/Przymusinski88}
\setcitation{}{}
\setcitation{}{}
\setcitation{}{}
\setcitation{}{}
\setcitation{}{}
\setcitation{}{}
\setcitation{}{}
\setcitation{}{}
\setcitation{}{}
\setcitation{}{}
\setcitation{}{}
\setcitation{}{}
\setcitation{}{}
\setcitation{}{}
\setcitation{}{}
\setcitation{}{}
\setcitation{}{}
\setcitation{}{}
\setcitation{}{}
\setcitation{}{}
\setcitation{}{}
\setcitation{}{}
\setcitation{}{}
  
\newcommand\refto[1]{%
      \getcitation{#1}}
      
\newcommand\mycite[1]{%
      \ifcsname mycommoncitation#1\endcsname%
   \cite{\getcitation{#1}}%
  \else%
    \cite{#1}%
  \fi%
}	
  
\usepackage{amsthm}
\usepackage{thmtools}

\declaretheorem[style=plain,	name=Theorem,		numberwithin=section]{thm}

\declaretheorem[style=plain,	name=Proposition,	numberlike=thm]{proposition}

\declaretheorem[style=plain,	name=Lemma,		numberlike=thm]{lemma}
\declaretheorem[style=plain,	name=Lemma,		numbered=no]   {lem*}

\declaretheorem[style=definition,	name=Definition,	numberlike=thm]{definition}

\declaretheorem[style=definition,	qed=$\blacktriangle$,	numberlike=thm]{example}

\declaretheorem[style=definition,	qed=$\blacktriangle$,	numbered=no]{ex*}

\declaretheorem[style=remark,	name=Notation,		numbered=no]{nota*}

\declaretheorem[style=remark,	name=Note,		numbered=no]{note*}

\ignore{

}
\setboolean{commentmargin}{false}

\renewcommand\voc{\m{\tau}}
\renewcommand\struct{\m{\mathcal{A}}}
\newcommand\pstruct{\m{\mathfrak{A}}}
\newcommand\pstructt{\m{\mathfrak{B}}}
\newcommand\dom{\m{A}}

\setdefaultenum{\bfseries(1)}{}{}{}
\newcommand\E{\m{E}}
\newcommand\hereisaproof[1]{}

\renewcommand{\emptyset}{\m{\varnothing}}
\newcommand{\domain}{\m{A}}
\newcommand{\bigtimes}{{\mathlarger{\mathlarger{\mathlarger{\mathlarger{\times}}}}}}

\newcommand\moduleof[1]{\m{\mathit{module}(#1)}}
\newcommand\solverof[1]{\m{\mathit{S}^P_{\mathit{p}}}}
\newcommand\gencheck[1]{\m{\mathit{S}^P_{\mathit{gc}}}}
\newcommand\canon[1]{\m{P_{\mathit{check}}^{#1}}}
\newcommand\checker[1]{\canon{#1}}
\newcommand\optimal[1]{\m{P_{\mathit{opt}}^{#1}}}

\newcommand{\mostprec}{\m{\mathfrak{I}}}
\newcommand\partitle[1]{\paragraph{#1.}}
\setboolean{nocomments}{true}
\renewcommand\hereisaproof[1]{#1}

\title{Propagators and Solvers for the Algebra of Modular Systems} 
\author{Bart Bogaerts\inst{1,2}\thanks{Bart Bogaerts is a postdoctoral fellow of the Research Foundation -- Flanders (FWO). This work is supported by the Finnish Center of Excellence in Computational
Inference Research (COIN) funded by the Academy of Finland (under
grant \#251170).}
\and Eugenia Ternovska\inst{3}
\and David Mitchell\inst{3} }

\institute{
\ignore{Helsinki Institute for Information Technology HIIT,}
Department of Computer Science, Aalto University, Espoo, Finland \and
KU Leuven, Department of Computer Science, Leuven, Belgium 
\and
   Computational Logic Laboratory, Simon Fraser University, Vancouver, Canada
}

\authorrunning{Bart Bogaerts et al.}

\titlerunning{Propagators and Solvers for the AMS}

\begin{document}
\maketitle
\begin{abstract}
Solving complex problems can involve non-trivial combinations of distinct knowledge bases and problem solvers. The Algebra of Modular Systems is a knowledge representation framework that provides a method for formally specifying such systems in purely semantic terms. Many practical systems based on expressive formalisms solve the model expansion task. In this paper, we construct a solver for the model expansion task for a complex modular system from an expression in the algebra and black-box propagators or solvers for the primitive modules. To this end, we define a general notion of propagators equipped with an explanation mechanism, an extension of the algebra to propagators, and a lazy conflict-driven learning algorithm. The result is a framework for seamlessly combining solving technology from different domains to produce a solver for a combined system.

\end{abstract}
\section{Introduction}

Complex artifacts are, of necessity, constructed by assembling simpler components.  Software systems 
use libraries of reusable components, and often access multiple remote services.   In this paper, we 
consider systems that can be formalized as solving the model expansion task for some class of finite 
structures.  A wide range of problem solving and query answering systems are so accounted for.
We present a method for automatically generating a solver for a complex system from a declarative 
definition of that system in terms of simpler modules, together with solvers for those modules.   
The work is motivated primarily by ``knowledge-intensive'' computing contexts, where the individual 
modules are defined in (possibly different) declarative languages, such as logical theories or logic programs, 
but can be applied anywhere the model expansion formalization can.

The Algebra of Modular Systems (AMS) \cite{frocos/TasharrofiT11,nmr/TasharrofiT14}, provides  a way to 
define a complex module in terms of a collection of other modules, in purely semantic terms.   Formally, 
each module in this algebra represents a class of structures, and a ``solver'' for the module solves the 
model expansion task for that class.   That is, a solver  for module ${\cal M}$ takes as input a structure 
${\cal A}$ for a part of the vocabulary of ${\cal M}$, and returns either a set of expansions of ${\cal A}$
that are in ${\cal M}$, or the empty set.  The operators of the algebra operate on classes of structures, 
essentially generalizing Codd's Relation Algebra \cite{cacm/Codd70} from tables to classes of structures. 

While the AMS provides a good account of how to define a module in terms of other modules, little work 
has been done on combining solvers (see e.g., \cite{lpnmr/MitchellT15}).  The purpose of this paper is to fill this gap.
To this end, we give general definitions of propagators and solvers in terms of classes of partial structures, 
and define algebras that correspond to the AMS, but operate on propagators rather than classes of structures.
The algebra provides a theoretical basis for the practical combination of solvers.

Intuitively, a propagator adds information to a partial structure without eliminating any solutions that 
extend that partial structure, while a solver searches through the set of more precise structures 
for a complete structure that is in the module. 
To formalize propagators and the operations of solvers, we use partial structures defined over 
the truth values \lunkn, \lfalse, \ltrue, \lincon (unknown, false, true and inconsistent, respectively),
with the ``precision order'' $\lunkn <_p \ltrue <_p \lincon$, $\lunkn <_p \lfalse<_p \lincon$.  Extending this 
order pointwise to structures, we obtain a complete lattice of partial structures ordered by precision.
The least structure in the lattice carries no information (all formulas are ``unknown''), while the 
greatest structure is inconsistent.  (For simplicity, in this paper we 
consider only finite vocabularies and finite structures.)

\partitle{Contributions}  Our main contributions are as follows.

\begin{compactenum}
\item  
We define a propagator for a module ${\cal M}$ to be an operator on the lattice of partial structures 
that is monotone (${\frak A} \geq_p  {\frak A}'$ implies $P({\frak A}) \geq_p P({\frak A}')$, and 
and information-preserving ($P({\frak A})\geq_p {\frak A}$).
This concept generalizes many uses of the term ``propagator'' in the literature.

\item 
We define an algebra of propagators that has the same operators and the same vocabulary as AMS, but the domain is 
propagators rather than modules.  The propagator algebra is analogous to the AMS in the 
following sense.  Each propagator $P$ is for a unique module, $\moduleof{P})$.  The function 
$\mathit{module}$ is a surjective homomorphism from the algebra of propagators to the AMS.
For example, if $\E_1, \E_2$ are modules, then 
$\pi_\delta (\sigma_{(Q\equiv R)} \E_1\times \E_2)$ is a compound module. It represents the class of structures \struct such that some structure $\struct'$ exists that coincides with \struct on $\delta$ and such that $\struct'$ satisfies both $\E_1$ and $\E_2$ and interprets $Q$ the same as $R$. 
If $P_1$ and $P_2$ are propagators for $\E_1$ and $\E_2$ respectively, then we show that in our extended version of the algebra, 
$\pi_\delta (\sigma_{(Q\equiv R)} P_1\times P_2)$
is a propagator for 
$\pi_\delta (\sigma_{(Q\equiv R)} \E_1\times \E_2).$

\item 
We show how solvers can be constructed from propagators, and vice versa.

\item We study \emph{complexity} of the combined propagators in terms of complexity of propagators for the individual modules and show that in general, our operations can increase complexity. 
This is useful, for example, to build a propagator for Quantified Boolean Formulas based on a propagator that (only) performs unit propagation for a propositional theory. We discuss how this can be done in Section \ref{sec:sts}. 

\item 
To model more interesting algorithms, we define {\em explaining propagators} to be propagators 
that return an ``explanation'' of a propagation in terms of simpler (explaining) propagators.  That is, 
an explaining propagator $P$ maps a partial structure to a more precise partial structure together 
with an ``explanation'' of that propagation in the form of an explaining propagator that is ``simpler''
than $P$.   These propagators generalizes lazy clause generation \mycite{LCG}, cutting plane 
generation \mycite{CuttingPlane} and counterexample-guided abstraction-refinement \mycite{CEGAR}.
This generalization shows that many techniques are actually instances of the same fundamental principles.

\item 
We extend the algebra of propagators to explaining propagators.

\item 
We show how to construct a conflict-driven learning solver for a module from an explaining 
propagator for the module.   The resulting algorithm is an abstract generalization of the 
conflict-driven clause learning (CDCL) algorithm for SAT, and other algorithms that can be 
found in the literature.   

\item 
We give several examples of techniques for applying the algebra in constructing solvers.

\end{compactenum}

Our formal framework results effectively in a paradigm where pieces of information (modules) are accompanied with implemented technology (propagators) and where composing solving technology is possible with the same ease as composing modules. 
The algorithms we propose are an important step towards practical applicability of the algebra of modular system.

\todo{
\rule{100pt}{3pt}
\begin{quote}

{\bf there are some useful bits of text here, but I don't know exactly where to put them yet.  Some might 
stay in intro/contributions, but others probably are more useful elsewhere.}

 Furthermore, it will, in the future, allow for very simple proofs of correctness for novel, similar, techniques. 
Indeed, it suffices to show that a new technique satisfies the general definitions from this paper to guarantee that it yields a correct (learning) solver.
We furthermore extend solvers based on these modular propagators with a \emph{conflict-analysis} method that generalizes resolution from conflict-driven clause learning \mycite{cdcl}.
%


 Our work extends SMT technology to arbitrary modules and to module combinations
| beyond conjunctive.

\end{quote}
\rule{100pt}{3pt}}

\partitle{Related Work}
The closest related work is research on technology integration. Examples include but are not limited
to \cite{lash/Gelfond08,lpnmr/BalducciniLS13,tplp/OstrowskiS12}. Combined solving is perhaps most developed in the \emph{SAT modulo theories (SMT)} community, where theory
propagations are tightly interleaved with satisfiability solving \cite{jacm/NieuwenhuisOT06,jsat/Sebastiani07}. 
The work we present in this paper differs from SMT in a couple of ways. 
In SMT, the problem associated with so-called \emph{theories} (modules in our terminology) is the satisfiability problem. This has several ramifications. One of them is that two different decidable theories cannot always be combined  into one decidable theories; this problem is known in the SMT community as the 
\emph{combination problem}; a large body of work has been devoted to the study of this problem, see for instance \cite{toplas/NelsonO79,frocos/TinelliH96,iandc/BaaderGT06,frocos/Fontaine09,cade/ChocronFR14,cade/ChocronFR15}.
In our approach on the other hand, the focus is not on the satisfiability problem, but on the model expansion problem, which is, from a complexity point of view, a simpler task than satisfiability checking \cite{lpar/KolokolovaLMT10}. As a consequence, we do not encounter problems such as the combination problem: any combination of propagators always yields a propagator. 
Instead, in our work, we focus on \emph{different ways to combine propagators}. In SMT, theories are always combined conjunctively.
For instance, SMT provides no support for negation (it is possible to state that the negation of an atom is entailed by a theory, but not that a theory is not satisfiable) or projection (projecting out certain variables in a given theory). This kind of combinations, generalizing the AMS to an algebra of propagators, results in a setting where propagators for simple modules (or ``theories'' in SMT terminology) can be combined into more complex propagators for the combined module. 
These combinations constitute the essence of our work.

Recently, Lierler and Truszczy{\'n}ski \cite{ai/LierlerT16} introduced a formalism with  compositions
(essentially, conjunctions) of modules given through solver-level inferences
 of the form $(M,l)$, where $M$ is a consistent set of
literals and $l$ is a literal not in $M$.
Such pairs are called inferences of the module.
Transition graphs for modules are constructed, with actions such as $\mathsf{Propagate}$, $\mathsf{Fail}$, $\mathsf{Backtrack}$, and $\mathsf{Decide}$.
Solvers based on the transition graph are determined by the
\emph{select-edge-to-follow} function (search strategy).
Solving templates are investigated for several formalisms, including SAT and ASP.
From individual transition graphs, such graphs are constructed for conjunctions
of modules, but more complex combinations of modules are not studied.

Combining propagators has been studied in constraint programming \cite{iclp/BrandY06,cp/GreenJ08,ai/JeffersonMNP10}. 
This research is often limited to a subset of the operations we consider here, for instance studying only conjunction  \cite{alp/Benhamou96,lopstr/AbdennadherF03}, disjunction \cite{wlp/MullerW95,ki/WurtzM96,cp/Lhomme03} or connectives from propositional logic  \cite{cpaior/Lhomme04,ijcai/BacchusW05}. In addition to these, we also consider selection and projection.
The objectives of the current paper are similar to those considered the CP community; however, there are some key differences. 
First of all, we generalized the theory of propagators from constraint programming to the AMS. 
It can be applied in principle to every logic with a model semantics. As such, it can serve as a formal basis to transfer the rich body of work from constraint programming to other fields, such as for instance (Integer) Linear Programming or Answer Set Programming.
Second, the traditional treatment of propagation in CP emphasizes tractability \cite{cp/GreenJ08}: the focus is on propagators that can be computed in polynomial time. While it is often important to constrain the complexity of the propagators, it can  be useful as well to allow for complexity increasing operations. In our framework, one of the operations (projection) increases complexity; as such, contrary to the propagators considered in constraint programming, it allows to construct propagators for compound modules for which membership checking is not polynomial. 
As explained above, this is useful to construct propagators for expressive logics such as QBF. 
Third, we equip our propagators with a \emph{learning mechanism} that generalizes for instance lazy clause generation \mycite{LCG} from constraint programming and a \emph{conflict analysis} mechanism that generalizes conflict-driven clause learning \mycite{cdcl}  from SAT \mycite{SAT}.

In real-world applications multiple (optimization) problems are often tightly intertwined. This has been argued intensively by Bonyadi et al.~\cite{cec/BonyadiMB13}, who designed the traveling thief problem as a prototype benchmarking problem of this kind. In the current state-of-the-art, such problems are tackled by special-purpose algorithms \cite{gecco/ChandW16,antsw/Wagner16}, illustrating the need for a principled approach to combining solving technology from different research fields.

\section{Modular Systems}\label{sec:modsys}



\partitle{Structures}
A (relational)\footnote{Without loss of generality, we restrict to relational vocabularies in this paper.} \emph{vocabulary} $\voc$ is a finite set of predicate symbols. A \emph{$\voc$-structure} $\struct$ consists of a domain \dom and an assignment of an $n$-ary relation $Q^\struct$ over \dom to all $n$-predicate symbols $Q\in\voc$.
A \emph{domain atom} is an expression of the form $Q(\ddd)$ with $Q\in\voc$, and \ddd a tuple of domain elements.
The value of a domain atom $Q(\ddd)$ in a structure $\struct$ (notation $Q(\ddd)^\struct$) is true ($\ltrue$) if $\ddd \in Q^\struct$ and false ($\lfalse$) otherwise. 
From now on, we assume that \domain is a fixed domain, shared by all structures. This assumption is not needed for the Algebra of Modular Systems in general, but it is convenient for the current paper since the input for the task we tackle (model expansion, see below) fixes the domain. 

A \emph{four-valued} $\voc$-structure \pstruct is an assignment $Q(\ddd)^\pstruct$ of a four-valued truth value (true ($\ltrue$), false ($\lfalse$), unknown ($\lunkn$) or inconsistent ($\lincon$)) to each domain atom over $\voc$.  If  $Q(\ddd)^\pstruct\in\{\ltrue,\lfalse,\lunkn\}$ for each $Q(\ddd)$, we call $\pstruct$ \emph{consistent}. If  $Q(\ddd)^\pstruct\in\{\ltrue,\lfalse\}$ for each $Q(\ddd)$, we call $\pstruct$ \emph{two-valued} and identify it with the corresponding structure.
A four-valued structures is sometimes also called a \emph{partial} structure, as it provides partial information about values of domain atoms. 

%
%
%
%
The precision order on truth values is induced by $\lunkn <_p \ltrue <_p \lincon$, $\lunkn <_p \lfalse <_p \lincon$. 
This order is pointwise extended to (four-valued) structures: $\pstruct <_p \pstruct'$ iff for all domain atoms $Q(\ddd)$, $Q(\ddd)^{\pstruct} <_p Q(\ddd)^{\pstruct'}$.
The set of all four-valued $\voc$-structures forms a complete lattice when equipped with the order $\leqp$. This means that every set $\mathcal{S}$ of (four-valued) structures has a greatest lower bound $\glb_{\leqp}(\mathcal{S})$ and a least upper bound $\lub_{\leqp}(\mathcal{S})$ in the precision order. 
Hence, there is a \emph{most precise} four-valued structure $\glb(\emptyset)$,  which we denote \mostprec; \mostprec is the most inconsistent structure: it maps all domain atoms to \lincon.

Four-valued structures are used to approximate structures. If \struct is a structure and \pstruct a partial structure, we say that \pstruct \emph{approximates} \struct if $\pstruct\leqp\struct$. 
Below, we use four-valued structures to represent a \emph{solver state}: certain domain atoms have been decided (they are mapped to $\ltrue$ or $\lfalse$), other domain atoms atoms have not yet been assigned a value (they are mapped to $\lunkn$), and certain domain atoms are involved in an inconsistency (they are mapped to $\lincon$). If a partial structure is inconsistent, it no longer approximates any structure. Solvers typically handle situations in which their state is inconsistent by backtracking.

If $Q(\ddd)$ is a domain atom and $\nu \in \{\ltrue,\lfalse,\lincon,\lunkn\}$, we use $\pstruct[Q(\ddd): \nu]$ for the (four-valued) structure equal to \pstruct except for interpreting $Q(\ddd)$ as $\nu$. We use $\pstruct[Q: Q^{\pstruct'}]$ for the four-valued structure equal to $\pstruct$ on all symbols except for $Q$ and equal to $Q$ on $\pstruct'$.  If $\delta\subseteq \voc$, we use $\pstruct|_\delta$ for the structure equal to $\pstruct$ on $\delta$ and mapping every other domain atom to $\lunkn$, i.e., $\pstruct|_\delta$  is the least precise structure that coincides with $\pstruct$ on $\delta$. 
%
%

\newcommand\mmodels[3]{\m{#1\models_{#2}#3}}
\newcommand\valuein[3]{\m{#1\in #3^#2}}
\newcommand\nmodels[3]{\m{#1\not\models_{#2}#3}}
\newcommand\vocof[1]{\m{\mathit{voc}(#1)}}
\newcommand\modints{\m{\mathcal{M}}}
\newcommand\modvals{\m{\mathcal{V}}}
\newcommand\cA{\m{\struct}}

\partitle{Modules}
Let $\tau_M =\{ M_1, M_2, \dots \}$ be a fixed vocabulary of {\em atomic module symbols} (often just referred to as \emph{atomic modules})
and let 
$\tau$ be a fixed  vocabulary.
  {\em Modules} are built by the grammar:
\begin{equation}\label{eq:algebra}
E::= \bot  \mid  M_i   \mid E\times E \mid
- E \mid \pi_{\delta} E \mid \sigma_{Q\equiv R} E.
\end{equation}
We call $\times$ \emph{product}, $-$ \emph{complement}, $\pi_\delta$ \emph{projection} onto $\delta$, and $\sigma_{Q\equiv R}$ \emph{selection}.
Modules that are not atomic are called {\em compound}.
Each atomic module symbol $M_i$  has an associated vocabulary $ \vocof{M_i} \subseteq \tau$.
The \emph{vocabulary} of a compound module is given by \begin{compactitem}\item$\vocof{\bot}=\tau$,  \item $\vocof{E_1\times E_2}=
\vocof{E_1} \cup \vocof{E_2}$, \item $\vocof{- E} =  \vocof{E}$, \item $\vocof{\pi_{\delta} E}=\delta$, and \item $\vocof{\sigma_\Theta E}=\vocof{E}$.\end{compactitem}


\partitle{Semantics}
Let ${\cal C} $ be the set of all  $\tau$-structures with domain $A$.
  Modules (atomic and compound) are interpreted by subsets of  ${\cal C} $.\footnote{If the assumption that $A$ is fixed is dropped, modules are {\em classes} of structures instead of sets of structures, but this generality is not needed
        in this paper.}
%
%
%
A \emph{module interpretation} {\cal I}  assigns to each atomic module $M_i\in \tau_M$ a set of $\tau$-structures
 such that any two $\tau$-structures $\cA_1$ and $\cA_2$ that coincide on $\vocof{M_i}$ satisfy
   $\cA_1 \in {\cal I}(M_i)$ iff  $\cA_2 \in {\cal I}(M_i)$.
The value of a modular expression $E$ in \I, denoted $\llbracket E\rrbracket^\I$, is defined as follows. 
$$
\begin{array}{l}

\llbracket \bot \rrbracket ^\I : =  \varnothing.\\
\llbracket M_i\rrbracket ^\I : = {\cal I} ( M_i).\\


\llbracket E_1 \times  E_2\rrbracket ^\I : = \llbracket E_1\rrbracket ^\I \cap  \llbracket E_2\rrbracket^\I.\\

\llbracket  -E \rrbracket ^\I  : =  {\cal C} -  \llbracket E\rrbracket ^\I.\\

\llbracket \pi_\delta(E) \rrbracket ^\I : = \{   \cA  \  \ |\   \exists {\cA}' \ (
       \cA' \in \llbracket E\rrbracket ^\I \mbox{ and }  {\cA}|_\delta={\cA}'|_\delta) \}.\\

\llbracket \sigma_{Q\equiv R} \E \rrbracket ^\I : = \{   \valuein{\struct}{\I}{ \E}  \mid  Q^{\cA} = R^{\cA} \}.\\


\end{array}
$$

\noindent We call $\cA$ a \emph{model} of $E$ in \I (denoted \mmodels{\struct}{\I}{E}) if $\struct\in\llbracket E\rrbracket^\I$.

\ignore{
From now on, we assume that a relational vocabulary $\voc$ and a set of symbols $M$ we call \emph{atomic modules} are fixed. 
Each atomic module has an associated vocabulary $\vocof{M}\subseteq \tau$. 
A \emph{module value} over vocabulary $\sigma\subseteq\tau$ is a class $\nu$ of $\tau$-interpretations  such that $\struct\in \nu$ iff $\struct'\in \nu$ for all interpretations $\struct$ and $\struct'$ with $\struct|_\sigma=\struct'|_{\sigma}$.
We use $\modvals$ to refer to the class of all module values. 
A \emph{module interpretation} \I associates to each atomic module $M$ a module value $\I(M)$ over $\vocof{M}$. 
We use $\modints$  to refer to the class of all module interpretations.
We say that $\struct$ is a \emph{model} of $M$ in \I (notation \mmodels{\struct}{\I}{M}) if $\struct\in \I(M)$.


Compound modules $\E$ are defined as follows from atomic modules $M$:
\[\E:= \top  \mid  M  \mid \E\times \E \mid  
- \E \mid \pi_{\tau'} \E \mid \sigma_\Theta \E \mid  \mu M. \E.\]
The \emph{vocabulary} of a compound module is given by $\vocof{\top}=\tau$, $\vocof{\E_1\times \E_2}=
\vocof{\E_1}\cup\vocof{\E_2}$, $\vocof{-\E} = \vocof{\E}$, $\vocof{\pi_{\tau'} \E}=\tau'$, $\vocof{\sigma_\Theta \E}=\vocof{\E}$ and $\vocof{\mu M. \E} = \vocof{\E}$. 
The following syntactic restrictions apply to the above expressions:
\begin{compactenum}
 \item For $\pi_{\tau'} \E$: $\tau'\subseteq \vocof{\E}$.
 \item For $\sigma_\Theta \E$: $\tau'\subseteq \tau$ and $\Theta$ is an expression of the form $P\equiv Q$ with $P,Q\in \vocof{\E}$.
 \item \label{item:three} For $\mu M. \E$: $\vocof{M} = \vocof{\E}$ and $M$ occurs only positively, i.e., in the scope of an even number of set complements ``$-$'', in $E$. 
\end{compactenum}

\noindent The \emph{value} of a compound module $\E$ in a module interpretation $\I$ (notation $\E^\I$) is a module value over $\vocof{\E}$, defined recursively. 
\begin{compactitem}
\item The \textbf{tautological module} is always satisfied: $\struct \in \top^\I$ for each structure $\struct$. 
\item For \textbf{atomic modules}, their value is stated in $\I$: $M^\I=\I(M)$.  
 \item \textbf{Product} (intersection) of two modules: $\valuein{\struct}{\I}{(\E_1\times  \E_2)}$ if $\valuein{\struct}{\I}{ \E_1}$ and $\valuein{\struct}{\I}{ \E_2}$ .
\item \textbf{Set complement} of a module: $\valuein{\struct}{\I}{- \E}$ if $\nmodels{\struct}{\I}{ \E}$
 \item \textbf{Projection} of a module: $\valuein{\struct}{\I}{ (\pi_\sigma \E_1)}$ if there exists a $\struct'$ with $\struct|_\sigma=\struct'|_\sigma$ such that $\valuein{\struct'}{\I}{ \E_1}$.
 \item \textbf{Selection} of a module $\E_1$ with respect to $\Theta$: $\valuein{\struct}{\I}{(\sigma_{P\equiv Q} \E_1)}$ if $\valuein{\struct}{\I}{ \E_1}$ and $P^\struct = Q^\struct$. 
 \item \textbf{Recursion} of a module $\E$ over atomic module symbol $M$ is defined as follows. If $M$ is an atomic module and $\E$ any module with $\vocof{M}=\vocof{E}$ and $\I$ is a module interpretation we define an operator 
 \[T_{M,\E,\I}:\modvals\to\modvals: \nu \mapsto \E^{\I[M: \nu]}.\]
 If $M$ and $\E$ satisfy the condition \textbf{(\ref{item:three})} above, this operator is monotone. 
 We then define $(\mu M.\E)^\I=\lfp T_{M,\E,\I}$.
 
 

\end{compactitem}
We call a \voc-structure \struct a \emph{model} of $\E$  in \I (notation $\mmodels{\struct}{\I}{\E}$) if $\valuein{\struct}{\I}{\E}$. 
We call a partial \voc-structure (i.e., a consistent four-valued structure) \pstruct a \emph{$3$-model} of $\E$ if $\struct\models \E$ for every \voc-structure $\struct$ more precise than \pstruct.
\todo{Needed?}
}

 In earlier papers \cite{frocos/TasharrofiT11,nmr/TasharrofiT14}, the algebra was presented slightly differently; here, we restrict to a  \emph{minimal syntax}; this is discussed in detail in Section \ref{sec:patterns}. 

\begin{example}\todo{RVWR1: one or two graphs would be helpful}
Let $\voc = \{\mathit{Edge},\mathit{Trans}\}$ be a vocabulary containing two binary predicates. 
Let \I be a model interpretation, $M_t$ a module with vocabulary $\voc$ such that $\struct\models_\I M_t$ if and only if $\mathit{Trans}^\struct$ is the transitive closure of $\mathit{Edge}^\struct$ and $M_f$ is a module with vocabulary $\{\mathit{Trans}\}$ such that $\struct\models_\I M_f$ if and only if $\mathit{Trans}^\struct$ is the full binary relation on $A$. 
Consider the following compound module 
\[E:= \pi_{\{\mathit{Edge}\}}(M_t \times (- M_f)).\]
Here, $E$ is a module with vocabulary $\{\mathit{Edge}\}$ such that $\struct\models_\I E$ if and only if $\mathit{Edge}^\struct$ is a disconnected graph: 
the module $M_t$ sets $\mathit{Trans}$ to be the transitive closure of $\mathit{Edge}$; the term $(-M_f)$ ensures that $\mathit{Trans}$ is not the full binary relation; these two modules are combined conjunctively and the result is projected onto $\{\mathit{Edge}\}$, i.e., the value of $\mathit{Trans}$ does not matter in the result.
\end{example}

From now on, we assume that a module interpretation $\I$ is given and fixed. Slightly abusing notation, we often omit the reference to \I and write, e.g., $\struct\models E$ instead of $\mmodels{\struct}{\I}{E}$. 

\partitle{Model expansion for modular systems}
The \emph{model expansion task} for modular systems is: given a (compound) module \E and a partial structure \pstruct, find a structure \struct (or: find all structures \struct) such that $\struct\geqp\pstruct$ and $\mmodels{\struct}{\I}\E$ (if one such exists).

Mitchell and Ternovska \cite{lpnmr/MitchellT15} have defined methods to apply the lazy clause generation (LCG) paradigm \cite{cp/FeydyS09} to solve the model expansion problem for modular systems. In particular, given propagators $P_i$ that explain their propagations by means of clauses  for atomic modules $M_i$, they show how to build an LCG-solver for modules of the form 
$\E = M_1 \times \dots \times M_n.$
%
In this paper, we generalize the above idea to a setting where $\E$ is an arbitrary (compound) module and the learning mechanism is not necessarily clause learning. 


\section{Propagators and Solvers}
\label{sec:propagators}

%
%
%
Now, we define a general notion of \emph{propagator} and show how propagators for atomic modules can be composed into propagators for compound modules. 
Intuitively, a propagator is a blackbox procedure that refines a partial (four-valued) structure by deriving consequences of a given module. 
%
%
%
%

\begin{definition}\label{def:propagator}
 A \emph{propagator} is a mapping $P$ from  partial structures to partial structures such that the following hold:
 \begin{compactitem}
  \item $P$ is \leqp-monotone:  whenever $\pstruct\geqp\pstruct'$, also $P(\pstruct)\geqp P(\pstruct')$. 
  \item $P$ is information-preserving: $P(\pstruct)\geqp \pstruct$ for each $\pstruct$.
 \end{compactitem}
\end{definition}

\begin{definition}\label{def:Epropagator}
Given a module $\E$, a propagator $P$ is an \emph{$\E$-propagator} if on two-valued structures, it coincides with $E$, i.e., whenever $\pstruct$ is two-valued, $P(\pstruct)=\pstruct$ iff $\pstruct\in \E$. 
\end{definition}

Note that if $\pstruct \not \in \E$, and $\pstruct$ is two-valued, an $E$-propagator maps \pstruct to an inconsistent partial structure since $P$ is information-preserving.
An $E$-propagator can never ``lose models of $E$'', as is formalised in the following lemma.
\begin{lemma}\label{lem:moreprecise}
 Let $P$ be an $\E$-propagator. If $\struct$ is a model of $\E$ and $\struct \geqp \pstruct$, then also $\struct \geqp P(\pstruct)$.
\end{lemma}
\hereisaproof{
\begin{proof}
 Follows from $\leqp$-monotonicity and the fact that $P(\struct)=\struct$ for two-valued structures \struct. 
\end{proof}}

\begin{example}\label{ex:ASPpropagators}
 Modern ASP solvers typically contain (at least) two propagators. One, which we call $P_{\mathit{UP}}^\PP$, performs unit propagation on the completion of the program $\PP$. The other, which we call $P_{\mathit{UFS}}^\PP$ performs unfounded set propagation; that is: it maps a partial structure \pstruct to 
 \[\lub_{\leqp}(\pstruct, \pstruct[p:\lfalse \mid p\in \mathit{lUFS}(\PP,\pstruct)]),\] where $\mathit{lUFS}(\PP,\pstruct)$ is the largest unfounded set of \PP with respect to \pstruct \cite{GelderRS91}. 
 It is easy to see that these two propagators are information-preserving and $\leqp$-monotone. 
 \end{example}

\begin{example}\label{ex:cp:propagator}
 In several constraint solvers that perform bounds reasoning, (finite-domain) integer variables are represented by a relational representation of their bounds: a variable $c$ is represented by a unary predicate $Q_{c\leq}$ with intended interpretation that $Q_{c\leq}(n)$ holds iff $c\leq n$. 
 Consider in this setting a propagator $P_{c\leq d}$ that enforces bounds consistency for the constraint $c\leq d$. 
 That is, $P_{c\leq d}$ maps a partial structure $\pstruct$ to a partial structure $\pstruct'$ such that for each $n$:
 \begin{itemize}
  \item if $ Q_{d\leq}(n)^\pstruct\geqp \ltrue$, then $Q_{c\leq}(n)^{\pstruct'}=\lub_{\leqp}(Q_{c\leq}(n)^\pstruct, Q_{d\leq}(n)^\pstruct)$,
  \item otherwise, $Q_{c\leq}(n)^{\pstruct'}=Q_{c\leq}(n)^{\pstruct}$
  \end{itemize}
  and similar equations for $Q_{d\leq}(n)$. Intuitively, these equations state that if $d\leq n$ holds in $\pstruct$, then $P_{c\leq d}$ also propagates that $c\leq n$ holds. 
 For instance, assume the domain $A=\{1,\dots, 100\}$ and that $\pstruct$ is such that
\begin{align*}  Q_{c\leq}(n)^\pstruct =\left\{\begin{array}{ll}\ltrue & \text{ if }n\geq 90 \\ \lunkn & \text{ if }90 > n \geq 10 \\ \lfalse & \text{ otherwise} \end{array}\right. &&
  Q_{d\leq}(n)^\pstruct =\left\{\begin{array}{ll}\ltrue & \text{ if }n\geq 80 \\ \lunkn & \text{ if }80 > n \geq 20 \\ \lfalse & \text{ otherwise} \end{array}\right.
 \end{align*}
 This structure encodes that the value of $c$ is in the interval $[10,90]$ and $d$ in the interval $[20,80]$. 
 In this case, $P_{c\leq d}$ propagates that also $c\leq 80$ without changing the value of $d$. 
 Formally:
\[  Q_{c\leq}(n)^{P_{c\leq d}(\pstruct)} =\left\{\begin{array}{ll}\ltrue & \text{ if }n\geq 80 \\ \lunkn & \text{ if }80 > n \geq 10 \\ \lfalse & \text{ otherwise} \end{array}\right.  \qquad 
 Q_{d\leq}(n)^{P_{c\leq d}(\pstruct)} =\ Q_{d\leq}(n)^\pstruct\qedhere
 \]
\end{example}

\partitle{Propagators and modules}
\begin{lemma}\label{lem:uniqueprop}
 If $P$ is a propagator, then there is a unique module $\E$ such that $P$ is an $\E$-propagator.
\end{lemma}
\hereisaproof{
\begin{proof}
 Uniqueness follows immediately from Definition \ref{def:Epropagator}. Existence follows from the fact that we can define the module $\E$ such that $\struct\models \E$ if and only if $P(\struct)=\struct$. 
\end{proof}}

Lemma \ref{lem:uniqueprop} gives rise to the following definition. 
\begin{definition}
If $P$ is a propagator, we define  $\moduleof{P}$ to be the unique module $\E$ such that $P$ is an $\E$-propagator.
\end{definition}

\begin{definition}
 If $\E$ is a module, the \emph{$\E$-checker} is the propagator \canon{\E} defined by:\footnote{Recall that \mostprec denotes the most precise (inconsistent) structure.}
 \[
  \canon{\E}: \pstruct \mapsto \left\{\begin{array}{ll}
  \pstruct & \text{if \pstruct is consistent but not two-valued}\\
  \pstruct & \text{if \pstruct is two-valued and $\pstruct\models\E$}\\
  \mostprec & \text{otherwise}
  \end{array}\right.
 \]
\end{definition}

\begin{lemma}\label{lem:canon}
 For each module $\E$, $\canon{\E}$ is an $\E$-propagator.
\end{lemma}
\hereisaproof{
\begin{proof}
That $\canon{\E}$ is a propagator follows from the fact that ${\mostprec}$ is more precise than any partial structure \pstruct. It follows immediately from the definition that \canon{\E} coincides with \E on two-valued structures. 
\end{proof}}

The $\E$-checker is the least precise $\E$-propagator, as the following proposition states. 
\begin{proposition}\label{prop:checker}
 For each $\E$-propagator $P$ and each consistent structure $\pstruct$, $P(\pstruct)\geqp \canon{\E}(\pstruct)$.
\end{proposition}

\partitle{Propagators and Solvers}
%


\begin{definition}
 Let \E be a module. An \emph{$\E$-solver} is a procedure that takes as input a four-valued structure \pstruct
  and whose output is the set $\mathcal{S}$ of all two-valued structures \struct with $\struct\models \E$ and $\struct\geqp\pstruct$. 
\end{definition}
Intuitively, a \emph{solver} is a procedure that performs model expansion for a given module. 
Propagators can be used to create solvers and vice versa. We first describe how to build a simple generate-and-check solver from a propagator. Afterwards, we provide an algorithm that uses the solver in a smarter way. In the next section, we discuss how to add a learning mechanism to this solver.

\begin{definition}\label{def:generate:check}
 Let $P$ be an $E$-propagator. We define an $E$-solver $\gencheck{P}$ as follows.\footnote{Here, $\mathit{gc}$ stands for Generate-and-Check.} 
  $\gencheck{P}$ takes as \textbf{input} a structure \pstruct. The \textbf{state} of $\gencheck{P}$ is a tuple $(\pstructt,\mathcal{S})$ of a structure and a set of two-valued structures $\mathcal{S}$; it is \textbf{initialised} as $(\pstruct,\emptyset)$. 
 $\gencheck{P}$ performs depth-first search on the search space of (four-valued) structures more precise than \pstruct. Choices consist of updating $\pstructt$ to $\pstructt[Q(\ddd): \ltrue]$ or to $\pstructt[Q(\ddd): \lfalse]$ for some domain atom with $Q(\ddd)^\pstructt=\lunkn$. 
 Whenever $\pstructt$ is two-valued, the solver checks whether $P(\pstructt)=\pstructt$. If yes, $\pstructt$ is added to $\mathcal{S}$. After encountering a two-valued structure, it backtracks over its last choice. 
 When the search space has been traversed, $\gencheck{P}$ \textbf{returns} $\mathcal{S}$. Pseudo-code for this algorithm can be found in Algorithm \ref{alg:gencheck}.
\end{definition}

\begin{algorithm}
  \caption{The solver $\gencheck{P}$
    \label{alg:gencheck}}
  \begin{algorithmic}[1]
    \Statex
     \Function{\gencheck{P}}{\pstruct}
      \Let{$(\pstructt,\mathcal{S})$}{$(\pstruct,\emptyset)$} 
      \While{ true }{}
       \If{$\pstructt$ is two-valued}
         \If{$P(\pstructt) = \pstructt$} 
          \Let{$S$}{$S\cup\{\pstructt\}$}
         \EndIf
         
         \If{the last choice updated $\pstructt'$ to $\pstructt'[Q(\ddd):\ltrue]$}\Let{\pstructt}{$\pstructt'[Q(\ddd):\lfalse]$}            
         \Else \hfill $\backslash\backslash$ {i.e., if there were no more choices}
          \State \Return S
         \EndIf
       \Else 
        \Let{choose \pstructt}{$\pstructt[Q(\ddd):\ltrue]$} for some $Q(\ddd)$ with $Q(\ddd)^\pstructt = \lunkn$
       \EndIf
      \EndWhile
    \EndFunction
  \end{algorithmic}
\end{algorithm}

\begin{definition}
Let $P$ be a propagator. We define the solver $\solverof{P}$ as follows.
 The solver $\solverof{P}$ extends $\gencheck{P}$ by updating \pstructt to $P(\pstructt)$ before each choice. If \pstructt is inconsistent, $\solverof{P}$ backtracks. Pseudocode for this solver can be found in Algorithm \ref{alg:solverof}.
\end{definition}

\begin{algorithm}
  \caption{The solver $\solverof{P}$
    \label{alg:solverof}}
  \begin{algorithmic}[1]
    \Statex
     \Function{\solverof{P}}{\pstruct}
      \Let{$(\pstructt,\mathcal{S})$}{$(\pstruct,\emptyset)$} 
      \While{ true }{}
       \Let{\pstructt}{P(\pstructt)}
       \If{$\pstructt$ is two-valued or inconsistent}
         \If{$P(\pstructt) = \pstructt$ and $\pstructt$ is two-valued} 
          \Let{$S$}{$S\cup\{\pstructt\}$}
         \EndIf
         
         \If{the last choice updated $\pstructt'$ to $\pstructt'[Q(\ddd):\ltrue]$}\Let{\pstructt}{$\pstructt'[Q(\ddd):\lfalse]$}            
         \Else \hfill $\backslash\backslash$ {i.e., if there were no more choices}
          \State \Return S
         \EndIf
       \Else 
        \Let{choose \pstructt}{$\pstructt[Q(\ddd):\ltrue]$} for some $Q(\ddd)$ with $Q(\ddd)^\pstructt = \lunkn$
       \EndIf
      \EndWhile
    \EndFunction
  \end{algorithmic}
\end{algorithm}

 
 \begin{proposition}
  If (the domain) $A$ is finite and $P$ is an $E$-propagator, then both $\gencheck{P}$ and $\solverof{P}$ are $E$-solvers.
 \end{proposition}
 \hereisaproof{
\begin{proof}[Sketch of the proof]
Finiteness of $A$ guarantees that depth-first search terminates. Correctness of $\gencheck{P}$ follows from the fact that $P$ is an $E$-propagator (since $\{\struct\mid P(\struct)=\struct\}= \{\struct\mid \struct\models E\}$). 

Correctness of $\solverof{P}$ follows from Lemma \ref{lem:moreprecise} which states that no models are lost by propagation. \end{proof}
}
 In the above proposition, the condition that $A$ is finite only serves to ensure termination of these two procedures that essentially traverse the entire space of structures more precise than \pstructt. 
 All the concepts defined in this paper, for instance propagators, and operations on propagators, can also be used in the context of an infinite domain. 
 
 

We now show how to construct a propagator from a solver. We call this propagator \emph{optimal}, since it always returns the most precise partial structure any propagator could return (cf.\ Proposition \ref{prop:opt}). In this sense, this propagator  performs skeptical reasoning. 
 \begin{definition}
  Let $S$ be an $E$-solver. We define an $E$-propagator $\optimal{S}:\pstruct\mapsto \glb_{\leqp} S(\pstruct).$
%
 \end{definition}
 
 Notice that if $S'$ is an $E$-solver as well, as a function, $\optimal{S}=\optimal{S'}$. However, we include $S$ in the notation since for practical purposes, we need a way to compute $\optimal{S}(\pstruct)$; for this, a call to $S$ is used. 
 \begin{proposition}
  If $S$ is an $E$-solver, $\optimal{S}$ is an $E$-propagator.
 \end{proposition}
 \hereisaproof{
 \begin{proof}
  We first show that $\optimal{S}$ is a propagator. 
  First, for each $\struct \in S(\pstruct)$, it holds that $\struct\geqp\pstruct$, hence also $\optimal{S}(\pstruct)\geqp\pstruct$. 
  Second, notice that whenever $\pstruct_1\geqp \pstruct_2$, $S(\pstruct_1)\subseteq S(\pstruct_2)$, hence $\optimal{S}(\pstruct_1)=\glb_{\leqp} S(\pstruct_1)\geqp \glb_{\leqp} S(\pstruct_2)=\optimal{S}(\pstruct_2)$. From these two properties, it follows that $\optimal{S}$ is indeed a propagator.
  
  The fact that it is also an $E$-propagator follows from the property that for two-valued $\struct$, $S(\struct)=\{\struct\}$ if $\struct\models \E$ and $S(\struct)=\emptyset$ otherwise. 
 \end{proof}}

 \begin{proposition}\label{prop:opt}
  Let $P$ be any $\E$-propagator and $S$ an \E-solver. For each structure $\pstruct$, it holds that 
  $\optimal{S}(\pstruct)\geqp P(\pstruct).$
 \end{proposition}
 \hereisaproof{
 \begin{proof}
 From Lemma \ref{lem:moreprecise}, we find that $P(\pstruct)\leqp \struct$ if $\struct\in S(\pstruct)$. Hence also $P(\pstruct)\leqp \glb_{\leqp} S(\pstruct)=\optimal{S}(\pstruct)$. 
 \end{proof}}
\partitle{Combining propagators}
First, we discuss how propagators for the same module can be combined. Afterwards, we extend the algebra of modular systems to propagators.

\begin{proposition}\label{prop:combine}
 Composition $P_1\circ P_2$  of two $E$-propagators is an $E$ propagator. In particular, if $P$ is an $E$-propagator, also $P^n$ (an abbreviation for $P\circ P\circ\dots\circ P$ ($n$ times)) is an $E$-propagator. 
 We use $P^\infty$ for $\lim_{\leqp} P^n = \lub_{\leqp}\{P^n\mid n\in \nat\}$.
\end{proposition}

\begin{proposition}
 If $P_1$ and $P_2$ are two $\E$-propagators, then $(P_1\circ P_2)(\pstruct)\geqp P_1(\pstruct)$ and $(P_1\circ P_2)(\pstruct)\geqp P_2(\pstruct)$ for each $\pstruct$. 
\end{proposition}
 Now, we show how checkers for compound expressions in the algebra can be built from propagators for atomic modules. 
 These checkers are sufficient for defining the algebra on propagators: if propagators for atomic modules are given, Proposition \ref{prop:various:checkers} provides us with the means to obtain a propagator for compound expressions. However, for practical purposes, we are often interested in better, i.e., more precise propagators. Hence, after this proposition, we investigate for which operations better propagators can be defined. 
\begin{definition}\label{def:various:checkers}
 Let $P$ be an $\E$-propagator, $P'$ an $\E'$-propagator and $\delta\subseteq\tau$. 
 We define following checkers (we only define their behaviour on two-valued structures since otherwise the behaviour of checkers is trivial):
 \begin{compactitem}
  \item $\checker{\bot}:\struct\mapsto\mostprec$
  \item $\checker{E\times E'}:\struct\mapsto \lub_{\leqp}\{P(\struct),P'(\struct)\}$
  \item $\checker{-E}:\struct\mapsto\left\{\begin{array}{ll}\struct&\text{if }P(\struct)=\mostprec\\\mostprec&\text{otherwise}\end{array}\right.$
  \item $\checker{\pi_\delta E}:\struct\mapsto\left\{\begin{array}{ll}\struct&\text{if }\solverof{P}(\struct|_{\delta})\neq\emptyset \\\mostprec&\text{otherwise}\end{array}\right.$
  \item $\checker{\sigma_{Q\equiv R} E}:\struct\mapsto\left\{\begin{array}{ll}\struct&\text{if }P(\struct)=\struct\text{ and } Q^\struct=R^\struct \\\mostprec&\text{otherwise}\end{array}\right.$
 \end{compactitem}
\end{definition}
\begin{proposition}\label{prop:various:checkers}
 The operations defined in Definition \ref{def:various:checkers} define checkers. Furthermore for each compound module $E''$, $\checker{E''}$ is an $E''$-checker.
\end{proposition}
\hereisaproof{
\begin{proof}[Sketch of the proof]
 Correctness for each of the above follows easily from the definition of the semantics of modular systems.
\end{proof}
}

Now, we present for several of the operations a better (more precise) propagator (compared to only checking). 

\begin{proposition}\label{prop:better}
  Let $P$ be an $\E$-propagator, $P'$ an $\E'$-propagator and $\delta$ a sub-vocabulary of $\tau$. 
We define the following operations:
\begin{compactitem}
  \item $P\times P':\pstruct\mapsto \lub_{\leqp}\{P(\pstruct),P'(\pstruct)\}$. 
  \item $ \pi_\delta P: \pstruct\mapsto  \left\{
  \begin{array}{l} 	
   \mostprec \qquad \text{if $\pstruct$ is inconsistent}\\
   \mostprec \qquad \text{if $\pstruct$ is two-valued on $\delta$ and $\solverof{P}(\pstruct|_{\delta})=\emptyset$}\\
   \lub_{\leqp}(P(\pstruct|_{\delta})|_\delta, \pstruct|_{\tau\setminus\delta}) \qquad  \text{otherwise.}  
  \end{array}\right.$
  \item $\sigma_{Q\equiv R}P: \pstruct\mapsto (P(\pstruct))[Q:L,R:L]$ where $L=\lub_{\leqp}(Q^{P(\pstruct)},R^{P(\pstruct)})$. 
\end{compactitem}
It holds that $P\times P'$ is an $E\times E'$-propagator, $\pi_\delta P$ is a $\pi_\delta E$ propagator and $\sigma_{Q\equiv R}$ is a $\sigma_{Q\equiv R}E$-propagator. 
\end{proposition}
\hereisaproof{
\begin{proof}
  We provide a proof for projection; the other operations are analogous. 
  
  We show that $\pi_\delta P$ is a \textbf{propagator}.
  
 First, for each four-valued structure \pstruct, $P(\pstruct|_\delta)\geqp \pstruct|_\delta$ since $P$ is a propagator, hence also $\pi_\delta P(\pstruct)|_\delta = P(\pstruct|_\delta)|_\delta\geqp \pstruct|_\delta$.  Furthermore, $\pi_\delta P(\pstruct)|_{\tau \setminus \delta} = \pstruct|_{\tau\setminus\delta}$. Combining these two yields that $\pi_\delta P(\pstruct)\geqp\pstruct$ and hence that $\pi_\delta P$ is indeed information preserving (the cases where $P(\pstruct)=\mostprec$ are trivial). 
 
 We show $\leqp$-monotonicity of $\pi_\delta P$. Assume $\pstruct_1\geqp\pstruct_2$. 
 If $\pstruct_2$ is inconsistent, then so is $\pstruct_1$ and thus $\pi_\delta P(\pstruct_1)=\pi_\delta P(\pstruct_2)$. 
 If $\pstruct_2$ is two-valued on $\delta$ and $\solverof{P}(\pstruct_1|_{\delta})=\emptyset$, then either $\pstruct_1$ is inconsistent, or $\pstruct_1|_{\delta}=\pstruct_2|_{\delta}$. In both cases, the result is trivial. 
 If $\pi_\delta P(\pstruct_1)=\mostprec$, the result is trivial as well, hence we can assume that both $\pstruct_1$ and $\pstruct_2$ fall in the ``otherwise'' category in the definition of $\pi_\delta P$. 
 The $\leqp$-monotonicity of $\pi_\delta P$ now follows from the fact that if $\pstruct_1\geqp\pstruct_2$ then also \begin{inparaenum}\item $\pstruct_1|_\delta\geqp\pstruct_2|_\delta$ and thus $P(\pstruct_1|_\delta)\geqp P(\pstruct_2|_\delta)$ and \item $\pstruct_1|_{\tau\setminus\delta}\geqp\pstruct_2|_{\tau\setminus\delta}$. \end{inparaenum} 
Hence, we conclude that $\pi_\delta P$ indeed defines a propagator. 
 
  Now, we show that $\pi_\delta P$ is a \textbf{\bf $\mathbf{\boldsymbol\pi_{\boldsymbol\delta} E}$-propagator}. Let  \struct be a two-valued structure.
 
 First suppose $\struct\models \pi_\delta \E$. In this case, there exists a two-valued $\struct'$ such that $\struct'\models \E$ and $\struct|_\delta = \struct'|_\delta$. Thus, $\solverof{P}(\pstruct|_{\delta})\neq \emptyset$ in this case. Also, in this case $P(\struct|_\delta)$ is consistent and thus $P(\struct|_\delta)|_\delta =\struct|_\delta$. We conclude that in this case indeed $\pi_\delta P(\struct)=\struct$. 
 
 
 Now suppose $\struct\not\models \pi_\delta\E$. In this case, there exists no structure $\struct'$ such that $\struct'|_\delta = \struct|_\delta$ and $\struct'\models \E$. Thus $\solverof{P}(\struct|_{\delta})=\emptyset$ and $\pi_\delta(\struct)$ is indeed inconsistent.  
\end{proof}
}

The intuitions in the above proposition are as follows. For $P\times P'$, $P$ computes 
consequences of $E$, while $P'$ computes consequences of $E'$, given an input structure \pstruct. The propagator $P\times P'$ combines the consequences found by both: it returns the least upper bound of $P(\pstruct)$ and $P'(\pstruct)$ in the precision order. That is, it returns the structure in which all domain literals derived by any of the two separate propagators hold (and nothing more). 
For projection $\pi_\sigma P$, in the two-valued case, the solver \solverof{P} is used to check whether $\struct\in\pi_\delta E$. For the three-valued case, $P$ is used to propagate on $\pstruct|_{\delta}$, i.e., using only the information about the projected vocabulary $\delta$. From this propagation, only the information that is propagated about $\delta$ is kept (this is $P(\pstruct|_{\delta})|_\delta$). Indeed $\pi_\delta E$ enforces no restrictions on symbols in $\tau\setminus \delta$.  Furthermore, we transfer all knowledge we previously had on symbols in $\tau\setminus \delta$ (this is some form of inertia); the resulting structure equals $P(\pstruct|_{\delta})$ on symbols in $\delta$ and equals $\pstruct$ on symbols in $\tau\setminus \delta$.
For selection $\sigma_{Q\equiv R} P$, propagation happens according to $P$. Afterwards, all propagations for $Q$ are also transferred to $R$ and vice versa. This is done by changing the interpretations of both $Q$ and $R$ to the least upper bound (in the precision order) of their interpretations in $P(\pstruct)$.

\begin{example}[Example \ref{ex:ASPpropagators} continued]\label{ex:asp:cont:1}
 We already mentioned that typical ASP solvers have two propagators $P_{\mathit{UP}}^\PP$ and $P_{UFS}^\PP$. The actual propagation is done according to $P_{ASP}^\PP=P_{\mathit{UP}}^\PP\times P_\mathit{UFS}^\PP$. Furthermore, typically, this propagation is executed until a fixed point is reached, hence the entire propagation is described by $\left(P^\PP_{ASP}\right)^\infty$. 
 
 Now let $M^\PP$ be the module such that $\struct\models M^\PP$ if and only if \struct is a stable model of \PP. 
 It is well-known that a structure is a stable model of \PP if and only if it is a model of the completion and it admits no non-trivial unfounded sets \cite{pods/SaccaZ90}. 
 From this, it follows that $M^\PP=M_{\mathit{UP}}^\PP \times M_\mathit{UFS}^\PP$, where $M_{\mathit{UP}}^\PP$ is a module such that $\struct \models M_{\mathit{UP}}^\PP$ iff $\struct$ is a model of the completion of $\PP$ and  $M_\mathit{UFS}^\PP$ is a module such that $\struct \models M_\mathit{UFS}^\PP$ iff $\struct$ admits no non-trivial unfounded sets with respect to $\PP$. It is easy to see that $P_{\mathit{UP}}^\PP$ and $P_\mathit{UFS}^\PP$ are $M_{\mathit{UP}}^\PP$- and $M_\mathit{UFS}^\PP$-propagators respectively. From this it follows by Propositions \ref{prop:better} and \ref{prop:combine} that
 $P_{ASP}^\PP$ and also $\left(P^\PP_{ASP}\right)^\infty$ are $M^\PP$-propagators. 
\end{example}

Up to this point, we have described three different ways to construct $E$-propagators: \optimal{S} is the most precise $E$-propagator if $S$ is an $E$-solver, Proposition \ref{prop:various:checkers} describes  how to build $E$-checkers (the least precise propagators) from propagators for subexpressions of $E$ and Proposition \ref{prop:better} illustrates how to build more precise propagators for compound expressions. However, \emph{precision} is not the only criterion for ``good'' propagators. In practice, we expect propagators to be efficiently computable. We now show that this is indeed the case for the propagators defined in Proposition \ref{prop:better}.
\begin{proposition}Assume $A$ is finite; furthermore assume access to an oracle that computes $P(\pstruct)$ and $P'(\pstruct)$. The following hold
\begin{compactitem}
\item  $(P\times P')(\pstruct)$ can be computed in polynomial time (in terms of the size of $A$),
\item $(\sigma_{Q\equiv R}P) (\pstruct)$ can be computed in polynomial time,
\item $(\pi_\delta P)(\pstruct)$ can be computed in nondeterministic polynomial time. 
\end{compactitem} \label{prop:complexity}
\end{proposition}
\hereisaproof{
\begin{proof}[Sketch of the proof]
The first two statements follow easily from the definitions. For instance $(P\times P')(\pstruct)$ is defined as $\lub_{\leqp}\{P(\pstruct),P'(\pstruct)\}$. Computing this least upper bound can be done by comparing $Q(\ddd)^{P(\pstruct}$ and $Q(\ddd)^{P'(\pstruct}$ for each domain atom $Q(\ddd)$. There are only polynomially many domain atoms. 

For the last statement, the complexity is dominated by a call to $\solverof{P}(\pstruct|_{\delta})$, which is essentially depth-first search. \end{proof}
}
Proposition \ref{prop:complexity} shows that product and selection do not increase complexity when compared to the complexity of the propagators they compose. However, the situation for projection is different. That is not surprising, since Tasharrofi and Ternovska \cite{hr/TasharrofiT15} already showed that the projection operation increases the complexity of the task of deciding whether a structure is a member of a given module or not. As such, Proposition \ref{prop:complexity}  shows that our propagators for compound expressions only increase complexity when dictated by the complexity of checking membership of the underlying module. 



\section{Explanations and Learning } 
\label{sec:learning}
%
%
%
%

In many different fields, propagators are defined that \emph{explain} their propagations in terms of simpler constructs. 
For instance in CDCL-based ASP solvers \cite{\refto{clasp},\refto{wasp},\refto{minisatid}}, the unfounded set propagator explains its propagation by means of clauses. Similar explanations are generated for complex constraints in constraint programming (this is the lazy clause generation paradigm \cite{\refto{lcg}}) and in SAT modulo theories \mycite{DPLLT}. The idea to generate clauses to explain complex constraints already exists for a long time, see e.g.\ \cite{aaai/Mitchell98}. In integer programming, the cutting plane method \cite{or/DantzigFJ54} is used to enforce a solution to be integer. In this methodology, when a (rational) solution is found, a cutting plane is learned that explains \emph{why} this particular solution should be rejected.
Similarly, in QBF solving, counterexample-guided abstraction-refinement (the CEGAR methodology) \cite{\refto{cegar},ai/JanotaKMC16,fmcad/RabeT15} starts from the idea to first solve a relaxed problem (an abstraction), and on-the-fly add explanations \emph{why} a certain solution to the relaxation is rejected. 
More QBF algorithms are based on some of learning information on the fly \cite{sat/RanjanTM04,ijcai/JanotaM15,cp/ZhangM02,date/GoultiaevaSB13}. 
De Cat et al.\ \cite{jair/CatDBS15} defined a methodology where complex formulas are grounded on-the-fly. This is a setting where inference made by complex formulas is explained in terms of simpler formulas (in this case, formulas with a lower quantification depth). 

%
The goal of this section is to extract the essential building blocks used in all of the aforementioned techniques to arrive at an abstract, algebraic, framework with solvers that can learn new constraints/propagators during search. 
We generalize the common idea underlying each of the above paradigms by adding explanations and learning to our abstract framework. We present a general notion of an \emph{explaining propagator} and define a method to turn such a propagator into a solver that learns from these explanations. 
An explaining propagator is a propagator that not only returns the partial structure that is the result of its propagations ($P(\pstruct)$), but also an explanation ($C(\pstruct)$). This explanation takes the form of a propagator itself. Depending on the application, explanations must have a specific form. For instance, for lazy clause generation, the explanation must be a (set of) clause(s); in integer linear programming, the explanation  must be a (set of) cutting plane(s).
In general, there are two conditions on the explanation. First, it should explain why the propagator made certain propagations: $C(\pstruct)$ should derive at least what $P$ derives in \pstruct. Second, the returned explanation should not be arbitrary, it should be a consequence of the module underlying the propagator $P$.
%
%
%
\newcommand\unexplained{\m{\lozenge}}
\begin{definition}\label{def:explprop}
An \emph{explaining propagator} is tuple $(P,C)$ where 
 $P$ is a propagator
 and 
 $C$ maps each partial structure either to \textsc{unexplained} (notation $\unexplained$) or to an explaining propagator $C(\pstruct) = (P', C')$ such that the following hold:
 \begin{compactenum}
     \item (explains propagation) $P(\pstruct) \leqp P'(\pstruct)$. 
     \item (soundness): $\moduleof{P'}\subseteq\moduleof{P}$
\end{compactenum}
%
\end{definition}
\begin{example}\label{ex:cuttingplane}
 Integer linear programs are often divided into two parts: some solver performs search using linear constraints. When a solution is found, a \emph{checker} checks whether this solution is integer-valued. If not, this checker propagates inconsistency and \emph{explains} this inconsistency by means of a cutting plane. This process fits in our general definition of \emph{explaining propagator}: a cutting plane can be seen itself as a propagator: during search it can propagate that its underlying constraint is violated	. This propagator \emph{explains} the inconsistency and is a consequence of the original problem (namely of the integrality constraint).
\end{example}
 
As can be seen, we allow an explaining propagator to \emph{not explain} certain propagations. For instance, whenever $P(\pstruct)=\pstruct$, nothing new is derived, hence there is nothing to explain. We say that $(P,C)$ \emph{explains propagation from $\pstruct$} if either $P(\pstruct)=\pstruct$ or $C(\pstruct)\neq\unexplained$. 
Each propagator $P$ as defined in Definition \ref{def:propagator} can be seen as an explaining propagator $(P,C_\unexplained)$, where $C_\unexplained$ maps each partial structure to \unexplained.

\begin{example}[Example \ref{ex:cp:propagator} continued]\label{ex:cp:explanation}
 For each natural number $n$ let $cl_n$ denote the clause $Q_{c\leq }(n)\lor\lnot Q_{d\leq }(n)$ and let $P_n$ denote the propagator that performs unit propagation on $cl_n$.  For each $\pstruct$, let $U_\pstruct$ denote the set of all $n$'s such that at least one literal from $cl_n$ is false in $\pstruct$. 
 Furthermore, let $C_{c\leq d}$ denote the mapping that maps each four-valued structure \pstruct to 
  \begin{align*} \left\{
  \begin{array}{ll}
  \unexplained &\text{if $P_{c\leq d}(\pstruct)=\pstruct$}\\
  \left(\bigtimes_{n \in U_\pstruct}
  P_n,C_\unexplained\right) &\text{otherwise},
 \end{array}\right.
\end{align*}
where $\bigtimes_{n \in U_\pstruct}
  P_n$ denotes the product of all $P_n$ with $n\in U_\pstruct$. 
In this case, $(P_{c\leq d}, C_{c\leq d})$ is an explaining propagator. It explains each propagation by means of a set of clauses (the product of propagators $P_n$ for individual clauses). This particular explanation is used for instance in MinisatID \mycite{CPsupport} and many other lazy clause generation CP systems. 
\end{example}

In general, using \emph{anything} as explanation is a bad idea: what we are hoping for is that a propagator explains its propagations in terms of \emph{simpler} propagators (where the definition of ``simple'' can vary from field to field). 
In order to generalize this idea, in what follows we assume that $\prec$ is a strict well-founded order on the set of all explaining propagators, where smaller propagators are considered ``simpler''. 
In the following definition, we also require that all propagations need to be explained, except for $\prec$-minimal propagators. 

\todo{a well-founded order $\prec$ s.t. the second condition in the definition below is satisfied always exists: there can never be ``learning loops'' given the inductive definitions of ``learning propagator''. However, the first condition in the following definition is not satisfied for that order}

\begin{definition}\label{def:respects}
We say that an explaining propagator $(P,C)$ \emph{respects} $\prec$ if 
\begin{compactitem}
 \item Whenever $C(\pstruct) = \unexplained$, $P(\pstruct)=\pstruct$ or $(P,C)$ is $\prec$-minimal,
 \item In all other cases, $C(\pstruct) \prec (P,C)$ and $C(\pstruct)$ respects $\prec$. 
\end{compactitem}
\end{definition}

\begin{example}\label{ex:lcg}
 For lazy clause generation, the order on propagators would be $(P,C) \prec (P',C')$ if $P$ performs unit propagation for a set of clauses and $P'$ does not. The conditions in Definition \ref{def:respects} guarantee that each non-clausal propagator explains its propagation in terms of these $\prec$-minimal propagators; i.e., in terms of clauses. 
\end{example}

\begin{example}
 When grounding lazily \mycite{lazygrounding}, one can consider propagators $P_\varphi$ that perform some form of propagation for a first-order formula $\varphi$. 
 A possible order $\prec$ is then: $(P_\varphi,C_\varphi)\prec (P_{\varphi'},C_{\varphi'})$ if $\varphi$ has strictly smaller quantification depth then $\varphi'$. 
 For instance, a propagator for a formula $\forall x: \exists y: \psi(x,y)$ can explain its propagations by means of a propagator for the formula $\exists y: \psi(d,y)$, where $d$ is an arbitrary domain element.
\end{example}


\newcommand{\lcgsolver}[1]{\m{\mathit{ls}^{#1}}}
\newcommand{\cdlsolver}[1]{\m{\mathit{cdl}^{#1}}}
\newcommand{\invalidateprop}[1]{\m{P^{#1}_{\mathit{inv}}}}
%

We now show how explaining propagators can be used to build solvers. 
\begin{definition}
\label{def:LearningSolver}
Let  $(P,C)$ be an explaining propagator that respects $\prec$. We define a \emph{learning solver} $\lcgsolver{(P,C)}$ as follows. 
The \textbf{input} of \lcgsolver{(P,C)} is a partial structure \pstruct. 
The \textbf{state} of \lcgsolver{(P,C)} 
 is a triple $(\PP,\pstructt,\mathcal{S})$ where 
 $\mathcal{P}$ is a set  of explaining propagators,
 \pstructt is a (four-valued) structure, and 
 $\mathcal{S}$ is a set  of (two-valued) structures.
The state is \textbf{initialised} as $(\{(P,C)\},\pstruct,\emptyset)$. 
The solver performs depth-first search on the structure \pstructt, where each choice point consists of assigning a value to a domain atom unknown in \pstructt. Before each choice point, until a fixed point is reached, \pstructt is updated to $P^*(\pstructt)$ and $\mathcal{P}$ to $\mathcal{P}\cup\{C^*(\pstructt)\}$, where $(P^*,C^*)$ is $\prec$-minimal among all elements of $\mathcal{P}$ that have $P^*(\pstructt)\neq\pstructt$. If no such element exists, no more propagation is possible and the solver makes another choice.
 Whenever \pstructt is inconsistent, the solver backtracks over its last choice. 
 If this search encounters a model (a two-valued structure $\struct$ with $\struct=P(\struct)$), it stores this model in $\mathcal{S}$ and adds $(\checker{-\{\struct\}},C_\unexplained)$ to $\mathcal{P}$.
 After the search space has been traversed (i.e., inconsistency is derived without any choice points left), it returns $\mathcal{S}$. Pseudocode for this solver can be found in Algorithm \ref{alg:learning}.
\end{definition}

\begin{algorithm}
  \caption{The solver $\lcgsolver{(P,C)}$
    \label{alg:learning}}
  \begin{algorithmic}[1]
    \Statex
     \Function{\solverof{P}}{\pstruct}
      \Let{$(\PP,\pstructt,\mathcal{S})$}{$(\{(P,C)\},\pstruct,\emptyset)$} 
      \While{ true }{}
       \While{At least one $(P^*,C^*)\in \mathcal{P}$ has    $P^*(\pstructt)\neq\pstructt$}   
        \State Let $(P^*,C^*)$ be a $\prec$-minimal propagator with $P^*(\pstructt)\neq\pstructt$ in $\mathcal{P}$ 
        \Let{\pstructt}{$P^*(\pstructt)$}
        \Let{$\mathcal{P}$}{$\mathcal{P}\cup\{C^*(\pstructt)\}$}
       \EndWhile
       \If{$\pstructt$ is two-valued or inconsistent}
         \If{$P(\pstructt) = \pstructt$ and $\pstructt$ is two-valued} 
          \Let{$S$}{$S\cup\{\pstructt\}$}
          \Let{$\mathcal{P}$}{$\mathcal{P}\cup \{(\checker{-\{\struct\}},C_\unexplained)\}$}
         \EndIf
         
         \If{the last choice updated $\pstruct'$ to $\pstruct'[Q(\ddd):\ltrue]$}\Let{\pstructt}{$\pstructt'[Q(\ddd):\lfalse]$}            
         \Else \hfill $\backslash\backslash$ {i.e., if there were no more choices}
          \State \Return S
         \EndIf
       \Else 
        \Let{choose \pstructt}{$\pstructt[Q(\ddd):\ltrue]$} for some $Q(\ddd)$ with $Q(\ddd)^\pstructt = \lunkn$
       \EndIf
      \EndWhile
    \EndFunction
  \end{algorithmic}
\end{algorithm}

\begin{example}[Examples \ref{ex:asp:cont:1} and \ref{ex:lcg} continued]
\label{ex:asp:cont:2}
Consider the order $\prec$ from Example \ref{ex:lcg}. 
Note that $P_{\mathit{UP}}^\PP$ performs unit propagation on a set of clauses and hence is $\prec$-minimal.
In many modern ASP solvers \cite{\refto{minisatid},\refto{clasp},\refto{wasp}}, $P_\mathit{UFS}^\PP$ explains its propagations in terms of clauses as well, resulting in an explanation mechanism $C_\mathit{UFS}^\PP$. These clauses are typically obtained by applying acyclicity algorithms to the dependency graph of the program; for details, see for instance \cite{phd/Marien09}. 
In $\lcgsolver{}$, the order $\prec$ is then used to prioritize unit propagation over unfoundedness propagation, conform modern ASP practices. 
\end{example}

\begin{proposition}
Assume $A$ is finite. If $(P,C)$ is an $E$-explaining propagator, then 
\lcgsolver{(P,C)} is an $E$-solver. 
\end{proposition}
\hereisaproof{
\begin{proof}[Sketch of the proof]
As before, termination follows from the fact that $A$ is finite. It follows from the second condition in Definition \ref{def:explprop} that during execution of \lcgsolver{(P,C)} , for all $P'\in \mathcal{P}$, it holds that $\moduleof{P'}\subseteq\moduleof{P}$, hence propagation is indeed correct for $E$.
\end{proof}
}

The next question that arises is: \textit{how can explaining propagators for individual modules be combined into explaining propagators for compound modules?} The answer is not always simple. As for regular propagators, each operation can be defined trivially. In Section \ref{sec:propagators}, being defined trivially meant simply defining the checker. In this case, additionally,  it means that for $C$ we take $C_\unexplained$. Below, we discuss some more interesting cases. 
We will use the following notations. If $\ddd$ is a tuple of domain elements, we use 
$P_{Q\equiv R}^{\ddd}$ for the propagator that maps each structure $\pstruct$ to a structure equal to $\pstruct$ except that it interprets $Q(\ddd)$ and $R(\ddd)$ both as $\lub_{\leqp}(Q(\ddd)^\pstruct,R(\ddd)^\pstruct)$. 
We use $P_{Q\equiv R}$ for the propagator $\bigtimes_{\{\ddd\in A^n\}} P_{Q\equiv R}^{\ddd}$ where $A$ is the domain and $n$ is the arity of $Q$ and $R$. 
Furthermore, we use $C_{Q\equiv R}$ for the mapping that sends \pstruct  to
 \begin{align*} \left\{
  \begin{array}{ll} 	
   \unexplained &\text{if $P_{Q \equiv R}(\pstruct) = \pstruct$}\\
   \left(\bigtimes_{\{\ddd\in A^n\mid Q(\ddd)^\pstruct\neq R(\ddd)^\pstruct\}} P_{Q\equiv R}^{\ddd},C_\unexplained\right) &\text{ otherwise}
  \end{array}\right.
  \end{align*}


\begin{definition}\label{def:explainingOperations}
 Assume $(P,C)$ and $(P',C')$ are explaining propagators. We define the following explaining propagators:
\begin{compactitem}
 \item \textbf{Product}  of explaining propagators: $(P,C)\times(P',C')=(P\times P', C\times C')$ where 
 \begin{align*} C\times C':\pstruct\mapsto \left\{
  \begin{array}{ll}
  C(\pstruct)\times C'(\pstruct)&\text{if both $(P,C)$ and $(P',C')$}\\&\text{explain propagation from \pstruct}\\
  \unexplained &\text{otherwise}
 \end{array}\right.
\end{align*}

 \item \textbf{Projection} of an explaining propagator: $\pi_\delta(P,C) = (\pi_\delta P, \pi_\delta C)$, where 
 \begin{align*} \pi_\delta C : \pstruct\mapsto \left\{
  \begin{array}{ll} 	
   \unexplained &\text{if $\pstruct$ is inconsistent or $C(\pstruct|_\delta)=\unexplained$}\\
   \unexplained &\text{if $\pstruct$ is two-valued on $\delta$ and $s(P)(\pstruct|_{\delta})=\emptyset$}\\
   \pi_\delta (C(\pstruct)) & \text{otherwise}
  \end{array}\right.
\end{align*}
 \item \textbf{Selection} of an explaining propagator: $\sigma_{Q\equiv R} (P,C) = (P,C)\times (P_{Q\equiv R}, C_{Q\equiv R})$.
 
 \end{compactitem}
 
\end{definition}

\begin{proposition}
 The mappings in Definition \ref{def:explainingOperations} indeed define explaining propagators.
\end{proposition}
\hereisaproof{
\begin{proof}
 The proof is similar for all cases. We only give the proof for projection. 
 The proof is by induction on the structure of $C$. First assume that $C=C_\unexplained$. 
 In this case, $\pi_\delta (P,C) = (\pi_\delta P, C_\unexplained)$, which is indeed an explaining propagator. 
 
 For the induction case, we can assume that for each $\pstruct$ with $C(\pstruct)\neq \unexplained$, $\pi_\delta C(\pstruct)$ is an explaining propagator. 
 We show that $\pi_\delta(P,C)$ is an explaining propagator. Choose some $\pstruct$ with $(\pi_\delta C)(\pstruct) \neq\unexplained$.
 Let $(P',C')$ denote $(\pi_\delta C)(\pstruct)$ and $(P'', C'')$ denote $C(\pstruct)$. From Definition \ref{def:explainingOperations}, we know that $P'=\pi_\delta P''$.
 
 First, we show that $\pi_\delta(P,C)$ explains propagation, i.e., that $\pi_\delta P(\pstruct)\leqp P'(\pstruct)$. We know that $P(\pstruct)\leqp P''(\pstruct)$ since $(P,C)$ is an explaining propagator. It follows immediately from the definition of $\pi_\delta P$ that also $\pi_\delta P(\pstruct)\leqp \pi_\delta P''(\pstruct)=P'(\pstruct)$. 
 
 We now show that $\pi_\delta(P,C)$ only derives consequences, i.e., that $\moduleof{P'}\models \moduleof{\pi_\delta P}$. We know that $\moduleof{P''}\models \moduleof{P}$. From the definition of the semantics of modular systems, it follows that then also $\pi_\delta \moduleof{P''}\models \pi_\delta\moduleof{P}$. Furthermore, from Proposition \ref{prop:better}, we know that $\pi_\delta \moduleof{P''} = \moduleof{\pi_\delta P''}=\moduleof{P'}$ and $\pi_\delta \moduleof{P} = \moduleof{\pi_\delta P}$. The result then follows. 
%
%
%
\end{proof}
}







\subsection*{A Conflict-Driven Learning Algorithm}

The CDCL algorithm for SAT lies at the heart of most modern SAT solvers, and also
many SMT solvers, ASP solvers, and others.   We now give an algorithm scheme that 
generalizes the CDCL algorithm to modular systems.

\begin{definition}\label{def:cdl}
The solver \cdlsolver{(P,C)} is obtained by modifying $\lcgsolver{(P,C)}$ as follows. 
Each time propagation leads to an inconsistent state, we update 
 $\pstructt, \mathcal{P} \gets \pstructt^{\prime}, \mathcal{P}\cup \{(P',C')\}$, where  $(\pstructt^{\prime}, (P',C')) =$ HandleConflict$(\pstructt,\mathcal{P})$, 
 and HandleConflict is a function such that 
\begin{compactenum}
\item $(P',C')$ is an explaining propagator that respects $\prec$,
\item $\pstruct \leq_p \pstructt^{\prime} <_p \pstructt$ 
(backjumping),
\item  $(\mathcal{P}\cup \{(P',C')\})(\pstructt^{\prime})  >_p \mathcal{P}(\pstructt^{\prime})$ 
(learning),
\item $\moduleof{\mathcal{P}} \subseteq \moduleof{P'}$ (consequence)
\end{compactenum}
After executing HandleConflict, it is optional to restart by 
re-setting $\pstructt$ to $\pstruct$.

Pseudocode for this solver can be found in Algorithm \ref{alg:cdls}.
\end{definition}

The intuition is that HandleConflict is some function that returns a state to backtrack to, and a new propagator to add to the set of propagators. This new propagator should, in the structure to which we backtrack, propagate something that was not propagated before. Thus, by analyzing the conflict, we obtain better information and avoid ending up in the same situation again.

\begin{algorithm}
  \caption{The solver $\lcgsolver{(P,C)}$
    \label{alg:cdls}}
  \begin{algorithmic}[1]
    \Statex
     \Function{\solverof{P}}{\pstruct}
      \Let{$(\PP,\pstructt,\mathcal{S})$}{$(\{(P,C)\},\pstruct,\emptyset)$} 
      \While{ true }{}
       \While{At least one $(P^*,C^*)\in \mathcal{P}$ has    $P^*(\pstructt)\neq\pstructt$}   
        \State Let $(P^*,C^*)$ be a $\prec$-minimal propagator with $P^*(\pstructt)\neq\pstructt$ in $\mathcal{P}$ 
        \Let{\pstructt}{$P^*(\pstructt)$}
        \Let{$\mathcal{P}$}{$\mathcal{P}\cup\{C^*(\pstructt)\}$}
       \EndWhile
       \If{$\pstructt$ is two-valued}
          \Let{$S$}{$S\cup\{\pstructt\}$}
          \Let{$\mathcal{P}$}{$\mathcal{P}\cup \{(\checker{-\{\struct\}},C_\unexplained)\}$}
         
       \ElsIf{$\pstructt$ is inconsistent}
        \If{No choices were made}
         \Return S
        \EndIf
        \Let{$(\pstructt^{\prime}, (P',C'))$}{ HandleConflict$(\pstructt,\mathcal{P})$}
        \Let{$(\pstructt, \mathcal{P})$}{$(\pstructt^{\prime}, \mathcal{P}\cup \{(P',C')\}$}
       \Else
        \Let{choose \pstructt}{$\pstructt[Q(\ddd):\ltrue]$} for some $Q(\ddd)$ with $Q(\ddd)^\pstructt = \lunkn$
       \EndIf
      \EndWhile
    \EndFunction
  \end{algorithmic}
\end{algorithm}

\begin{proposition}
Assume $A$ is finite.
If $(P_E,C_E)$ is an $E$-explaining propagator that respects $\prec$, then \cdlsolver{(P_E,C_E)} is an $E$-solver. 
\end{proposition}\hereisaproof{
\begin{proof}[Sketch of the proof]
Correctness of $\cdlsolver{(P_E,C_E)}$ follows from correctness of $\lcgsolver{(P,C)}$ combined with the fourth condition for HandleConflict in Definition \ref{def:cdl}. 
The hardest thing to prove is termination of this algorithm in case restarts are involved. 
It can be seen that this algorithm terminates by the fact that after each conflict, by the third condition in HandleConflict, for at least one partial structure (namely for $\pstructt^{\prime}$),  strictly more is propagated by  $(\mathcal{P}\cup \{(P',C')\})$ than by $\mathcal{P}$. Since the number of propagators is finite, there cannot be an infinite such sequence, hence only a finite number of conflicts can occur. 
\end{proof}}

The purpose of HandleConflct is to perform a conflict analysis analogous to that in 
standard CDCL. This procedure can be anything; in practice, it will depend on the 
form $\prec$-minimal propagators take and on the proof system used for these minimal propagators.
Below, we present a sufficient restriction on $\prec$-minimal propagators to ensure that a procedure HandleConflict exists. 

\begin{proposition}
 Suppose that there exists a function $F$ that takes as arguments two $\prec$-minimal explaining propagators 
$(P_1,C_1)$ and $(P_2,C_2)$ that respect $\prec$, and a partial structure $\pstructt$, and returns an explaining 
propagator $(P,C)$ that respects $\prec$, such that the following hold. 
\begin{quote}
If $\pstruct <_p P_1(\pstructt) <_p P_2(P_1(\pstructt))$ and   
$\pstructt <_p P_2(\pstructt) <_p P_2(P_1(\pstructt))$, then 
$ \moduleof{P_1}\times \moduleof{P_2}  \subseteq \moduleof{P}$ 
and there exists a structure $\pstructt'\leqp \pstructt$ such that $P(\pstructt')>_p P_2(P_1(\pstructt'))$. 
\end{quote}
In that case, a procedure HandleConflict that satisfies the restrictions in Definition \ref{def:cdl} exists. 
\end{proposition}
\hereisaproof{
\begin{proof}[Sketch of the proof]
The idea is that it suffices to be able to combine $\prec$-minimal propagators since all propagations can (by iterated calls to the explanation mechanism) be explained in terms of these propagators. Furthermore, the above condition can be applied iteratively to combine more than two $\prec$-minimal propagators.
\end{proof}
}

The intuition for $F$ is that it effectively analyses the source of a 
conflict found by a sequence of propagations.
We want to be able to determine 
a minimum collection of points in the partial structure relevant to the conflict.
For this, it suffices that we can take two $\prec$-minimal propagators and ``resolve'' them to obtain one with stronger propagation power. 
Observe that, if we assume that all $\prec$-minimal 
propagators have a representation as clauses, this function can be implemented by means of the standard 
resolution used in CDCL conflict analysis process. In general, other resolution mechanisms might be used. The chosen implementation for $F$ essentially determines the \emph{proof system} that will be used in the solver. 

Iterated applications of $F$, starting from the last two propagators that changed state and working back to earlier propagators allow us to 
 handle conflicts. 
Note that $F$ is  only defined on $\prec$-minimal propagators. However, the explanation mechanism in explaining propagators allows us to always reduce propagators to $\prec$-minimal propagators by means of calling the explanation method until a minimal propagator is obtained (this is possible since $\prec$ is a well-founded order). 

\ignore{
We now describe one possible HandleConfict implementation.  
(Better ones may be possible for certain classes of propagators.)   

\begin{algorithm}
The arguments are $\pstructt$ and $\mathcal{P}$.  Assume steps in construction 
of $\pstructt$ are indexed, with each decision and each application of a propagator being a step.
For each domain atom $A$, we let $I(A)$ denote the first step where $A$ was assigned a value, 
and $R(A)$ denote the ``reason'' for $A$, which is ${\bf D}$ if  $A$ was a decision and the 
(first) propagator $P \in \mathcal{P}$ if $P$ to give $A$ a value otherwise. 
The reason for an inconsistent atom will be a pair to explain both the truth and falsity.
Let $ld$ be the index of the last decision, so that every atom $A$ set after that last decision 
has $I(A) > ld$.   

Now, let $C$ be an atom that is inconsistent in $\pstructt$, and assume that $C$ was 
set true before being set false. (The other case is symmetric.)

 and construct
a propagator $P_T$, explaining why $C$ is true, as follows.  If the reason for $C$ being 
true is ${\bf D}$, then let $P_T$ be the identity.  Otherwise, let $S = \{C\}$, and repeat 
the following steps until no atom $A \in S$ has $I(A) > ld$.  
Let $X$ be the set of atoms in $S$ with maximal index $i$.  Find the least partial structure 
$\pstructt^{\prime} \leq_p \pstructt$ such every atom of $X$ is known in 
$R(A)(\pstructt^{\prime})$, and let $Y$ be the set of atoms known in $\pstructt^{\prime}$.   
Set $S$ to $(S - X) \cup Y$.  When the iteration terminates, set $P_T$ to $A2(R(A),P_T)$.
Construct $P_F$ similarly, and set $P$ to $P_T \times P_F$.
Let $\pstructt^{\prime}$ be the most $\pstruct_i$ which is most  structure less precise than $\pstructt$ with 
$P(\pstruct^{\prime})$ inconsistent.

 \end{algorithm}
 }

%
%


\section{Modular patterns}\label{sec:patterns}
Sometimes, defining a propagator compositionally does not yield the best result. We identify three patterns for which we can define a better (more precise) propagator by exploiting a global structure. 
The first two optimizations consist of direct implementations for propagators for expressions in the algebra of modular systems that are not in the minimal syntax (for details, see, e.g., \cite{nmr/TasharrofiT14}). 
The third optimization is based on techniques that were recently used to nest different SAT solvers to obtain a QBF solver. 

\partitle{Disjunction of Modules}
The \emph{disjunction} of two modules is defined as $E_1+E_2 = -(-E_1\times -E_2)$. 
Such an operation can be used for instance in the context of a web-shop offering different shipping option. If the constraints related to each shipping option are specified by the shipping company itself, possibly in different languages, say in modules $M_{S_i}$ for different shipping companies and the desires of the user are specified in a module $M_U$, then $M_U\times (M_{S_1} + \dots + M_{S_n})$ represents a module in which the users desires are satisfied by at least one shipping company. This can be used to see if the web shop can satisfy the request or not. 
\begin{definition}
 Let $P_1$ and $P_2$ be an $E_1$-, respectively $E_2$-, propagator. We define a propagator 
 \[
  P_1+P_2: \pstruct\mapsto \glb_{\leqp}(P_1(\pstruct),P_2(\pstruct)). 
 \]
 \end{definition}
 Intuitively, this propagator only propagates what holds in both $P_1(\pstruct)$ and $P_2(\pstruct)$. 
 As such, it indeed only derives consequences of the disjunction.
 
 \begin{proposition}
 The following hold:
 \begin{compactitem}
  \item $P_1+P_2$ is a $-(-E_1\times -E_2)$ propagator
  \item For each consistent partial structure \pstruct, \[(P_1+P_2)(\pstruct) \geqp -(-P_1\times -P_2)(\pstruct).\] 
 \end{compactitem}
\end{proposition}
\hereisaproof{
\begin{proof}[Sketch of the proof]
It is easy to see that $\struct \models -(-E_1\times -E_2)$ iff $\struct \models E_1$ or $\struct\models E_2$. 
The first point now follows directly from the definition of $P_1+P_2$. 
The second point follows from Proposition \ref{prop:checker} since $-(-P_1\times -P_2)$ is a checker. 
\end{proof}
}

\partitle{Extended selection}
It is also possible to allow expressions of the form $\sigma_\Theta E$ where $\Theta$ consists of expressions of the form $Q\equiv R$ or $Q\not \equiv R$ and propositional connectives applied to them (for semantics, see, e.g., \cite{nmr/TasharrofiT14}). Each such expression can be rewritten to the minimal syntax used in this paper, for instance $\sigma_{P\not\equiv Q} E$ is equivalent to $E\times - \sigma_{P\equiv Q} E.$
By taking an entire such formula into account at once, more precise reasoning is possible. 
\begin{proposition}\label{prop:equality}
Let $P$ be a $\sigma_\Theta E$-propagator with $\Theta$ an expression as above. 
If $\Theta\models Q\equiv R$, then $\sigma_{Q\equiv R}P$ is also a $\sigma_\Theta E$ propagator and for all partial structures \pstruct, $(\sigma_{Q\equiv R}P)(\pstruct) \geqp P(\pstruct)$. 
\end{proposition}
\hereisaproof{
\begin{proof}
Follows directly from the fact that in this case $\sigma_{Q\equiv R}(\sigma_{\Theta}E) = \sigma_{\Theta}E$. \qedhere \end{proof}
}

Proposition \ref{prop:equality} states that we can use (symbolic) equality reasoning on $\Theta$ to derive more consequences. The following example illustrates the extra propagation power Proposition \ref{prop:equality} yields. 
\begin{example}
Consider the module 
 $
 \sigma_{(Q\not\equiv R \lor R \equiv U)\land (Q \equiv R)} E.
 $
 It is easy to see that $R\equiv U$ is a consequence of the selection expression in this  module. As such, Proposition \ref{prop:equality} guarantees that we are allowed to improve propagators, to also propagate equality between $Q$ and $R$.  
\end{example}

\newcommand\structt{\mathcal{B}}
\partitle{Improved Negation}
\label{sec:sts}
Janhunen et al.\ \cite{aaai/JanhunenTT16} recently defined a solver that combines two SAT solvers. The essence of their algorithm can be translated into our theory as follows. 
Let $\tau$ and $\delta$ be vocabularies, $E$ a $\tau\cup\delta$-module and $S$ an $E$-solver. Assume that there is a procedure \texttt{Explain}
such that for each two-valued $\tau$-structure \struct such that $S(\struct)\neq \emptyset$, $\pstruct=\texttt{Explain}(\struct)$ is a partial $\tau$-structure such that 
\begin{compactitem}
 \item $\pstruct\leqp \struct$
 \item For each two-valued $\tau$ structure $\structt\geqp \pstruct$,  $S(\structt)\neq\emptyset$. 
\end{compactitem}
Thus, \texttt{Explain} explains \emph{why} a certain module is satisfiable. 
Given this, Janhunen et al.\  defined an explaining propagator $P$ for 
$- \pi_\tau E$. 
If $\pstruct$ is two-valued on $\tau$, $P$ calls $S(\pstruct|_\tau)$. If the result of this call is not empty, it propagates a conflict and generates an explanation using \texttt{Explain}$(\pstruct|_\tau)$; this explanation is a propagator that performs unit propagation for a clause, the negation of the returned partial interpretation (see \cite{aaai/JanhunenTT16} for details). 
Otherwise, $P$ maps $\pstruct$ to itself. 


This idea has been generalized to work for arbitrary QBF formulas \cite{bnp/BogaertsJT16}. 
It forms the essence of many SAT-based QBF algorithms  \cite{sat/RanjanTM04,cp/ZhangM02,date/GoultiaevaSB13,ijcai/JanotaM15}. 
Janhunen et al.\ \cite{aaai/JanhunenTT16}  improved this method by introducing a notion of an \emph{underapproximation}. That is, instead of using an $E$-solver $S$, they 
use an $\bar E$-solver $\bar S$, where $\bar E$ is some module derived from $E$. 
This allows them to run $\bar S$ \emph{before} $\pstruct$ is two-valued on $\tau$. The module $\bar E$ is constructed in such a way that from runs of $\bar S$, a lot of information can already be concluded without having a two-valued $\pstruct$.  Researching how these underapproximations generalize to modular systems is a topic for future work.   

%
%
%
%
%
%

\section{Conclusion and Future Work}
In this paper, we defined general notions of \emph{solvers} and \emph{propagators} for modular systems.
We extended the algebra of modular systems to \emph{modular propagators} and showed how to build solvers from propagators and vice versa. 
We argued that our notion of propagator generalizes notions from various domains.
Furthermore, we added a notion of \emph{explanations} to  propagators. These explanations generalize concepts from answer set programming, constraint programming, linear programming and more. 
We used these explanations to build \emph{learning solvers} and discussed how learning solvers can be extended with a conflict analysis method, effectively resulting in a generalization of CDCL to other learning mechanisms and hence also to other proof systems. 
Finally, we discussed several patterns of modular expressions for which more precise propagation is possible than what would be obtained by creating the propagators following the compositional rules. 

The main contribution of the paper is that we provide an abstract account of propagators, solvers, explanations, learning and conflict analysis, resulting in a theory that generalizes many existing algorithms and allows integration of technology of different fields.
Our theory allows one to build \emph{actual solvers} for modular systems and hence provides an important foundation for the practical usability of modular systems.

Several topics for future work remain. While the current theory provides a strong foundation, an \emph{implementation} is still needed to achieve practical usability. We intend to research more \emph{patterns} for which improved propagation is possible, and generalize the aforementioned \emph{underapproximations} to our framework. The Algebra of Modular Systems has been extended with a \emph{recursion} operator (see for instance \cite{nmr/Ternovska16}); the question ``What are good propagators for this operator?'' remains an open challenge.
\bibliographystyle{plain}
\bibliography{idp-latex/krrlib,refs-for-this-paper}


\end{document}